\documentclass[conference]{IEEEtran}

\IEEEoverridecommandlockouts
\usepackage{cite}
\usepackage{amsmath,amssymb,amsfonts}
\usepackage{algorithm}
\usepackage{algorithmic}
\usepackage{textcomp}

\usepackage{graphicx}
\usepackage{amsmath}
\usepackage{amsthm}
\usepackage{booktabs}
\usepackage{xcolor}
\usepackage[switch]{lineno}
\usepackage{subcaption}
\usepackage{amssymb}  
\usepackage{multirow}
\usepackage{hyperref}

\def\BibTeX{{\rm B\kern-.05em{\sc i\kern-.025em b}\kern-.08em
    T\kern-.1667em\lower.7ex\hbox{E}\kern-.125emX}}

\newcommand{\stab}{\rule{0pt}{8pt}\\[-2.0ex]}
\newcommand{\eat}[1]{}
\newcommand{\eg}{{\it e.g.}, }
\newcommand{\ie}{{\it i.e.}, }

\newcommand{\LLMComp}{{\sc LLMComp}}

\newcommand{\myNum}[1]{(\emph{#1})}

\def\RS{\textsc{RedSea}}
\def\ERA{\textsc{Era}}

\title{\LLMComp: A Language Modeling Paradigm for Error-Bounded Scientific Data Compression (Technical Report)}

\author{%
\parbox{\textwidth}{\centering
  \begin{tabular}{c@{\hspace{4em}}c}
    \parbox{.35\textwidth}{\centering
      Guozhong Li\\
      KAUST\\
      Thuwal, Saudi Arabia\\
      guozhong.li@kaust.edu.sa}
    &
    \parbox{.35\textwidth}{\centering
      Muhannad Alhumaidi\\
      KAUST\\
      Thuwal, Saudi Arabia\\
      muhannad.humaidi@kaust.edu.sa}
  \end{tabular}
}%

\\[2.5ex] %

\parbox{\textwidth}{\centering
  \begin{tabular}{c@{\hspace{4em}}c}
    \parbox{.35\textwidth}{\centering
      Spiros Skiadopoulos\\
      University of Peloponnese\\
      Tripoli, Greece\\
      spiros@uop.gr}
    &
    \parbox{.35\textwidth}{\centering
      Panos Kalnis\\
      KAUST\\
      Thuwal, Saudi Arabia\\
      panos.kalnis@kaust.edu.sa}
  \end{tabular}
}%
}

\begin{document}

\maketitle

\begin{abstract}
    The rapid growth of high-resolution scientific simulations and observation systems is generating massive spatiotemporal datasets, making efficient, error-bounded compression increasingly important. 
    Meanwhile, decoder-only large language models (LLMs) have demonstrated remarkable capabilities in modeling complex sequential data.
    In this paper, we propose \LLMComp, a novel lossy compression paradigm that leverages decoder-only large LLMs to model scientific data.
    \LLMComp~first quantizes 3D fields into discrete tokens, arranges them via Z-order curves to preserve locality, and applies coverage-guided sampling to enhance training efficiency.
    An autoregressive transformer is then trained with spatial-temporal embeddings to model token transitions.
    During compression, the model performs top-$k$ prediction, storing only rank indices and fallback corrections to ensure strict error bounds.
    Experiments on multiple reanalysis datasets show that \LLMComp~consistently outperforms state-of-the-art compressors, achieving up to 30\% higher compression ratios under strict error bounds.
    These results highlight the potential of LLMs as general-purpose compressors for high-fidelity scientific data.
\end{abstract}


\section{Introduction}\label{sec:introduction}
The rapid growth of scientific simulations and environmental sensing systems~\cite{kay2015community} has led to an explosion of high-resolution scientific data\footnote{Scientific data spans many types, \eg text and image; this work focuses specifically on \emph{numerical} data.}. Applications in climate modeling~\cite{hoteit-RSRA2018,hoteit-RSRA2022} and atmospheric science~\cite{hersbach2020era5} routinely produce multi-terabyte datasets comprising temperature, pressure, humidity, and other physical variables across space and time. Efficient compression of such data, while preserving fidelity required for downstream analysis~\cite{yan2022sensor,li2025lesax}, has become a critical challenge in modern scientific computing. 

To address this challenge, traditional \emph{error-bounded} lossy compressors such as SZ~\cite{liu2021exploring,liu2023srn,liu2023high}, SPERR~\cite{li2023lossy}, and ZFP~\cite{lindstrom2014fixed} employ fixed local predictors and transform coding. These methods are effective in smooth regions where local predictors suffice, but fail to capture nonlinear dynamics and long-range dependencies typical of scientific fields~\cite{li2025graphcomp}. Moreover, their handcrafted heuristics often generalize poorly across heterogeneous variables and spatial resolutions. In contrast, \emph{error-unbounded} methods~\cite{huang2023compressing,han2024cra5} achieve higher compression by relaxing fidelity constraints, but fundamentally lack the point-wise error control required in scientific applications. This trade-off motivates the exploration of learning-based compressors capable of modeling complex structures while ensuring error-bounded guarantees.

Existing approaches for scientific data compression, especially in climate modeling, have predominantly employed encoder-decoder architectures trained for continuous regression~\cite{price2025probabilistic,mirowski2024neural,lam2023learning}. In parallel, recent advances in large language models (LLMs), notably decoder-only transformers such as GPT~\cite{radford2018improving}, Qwen~\cite{bai2023qwen}, and LLaMA~\cite{touvron2023llama}, have demonstrated exceptional capabilities in capturing long-range dependencies through next-token prediction across language and multimodal domains. Despite this success, their potential for modeling numerical scientific data, particularly for error-bounded compression, remains largely unexplored.

\begin{figure}[t]
    \centering
    \begin{subfigure}[b]{0.6\linewidth}
        \centering
        \includegraphics[width=\linewidth]{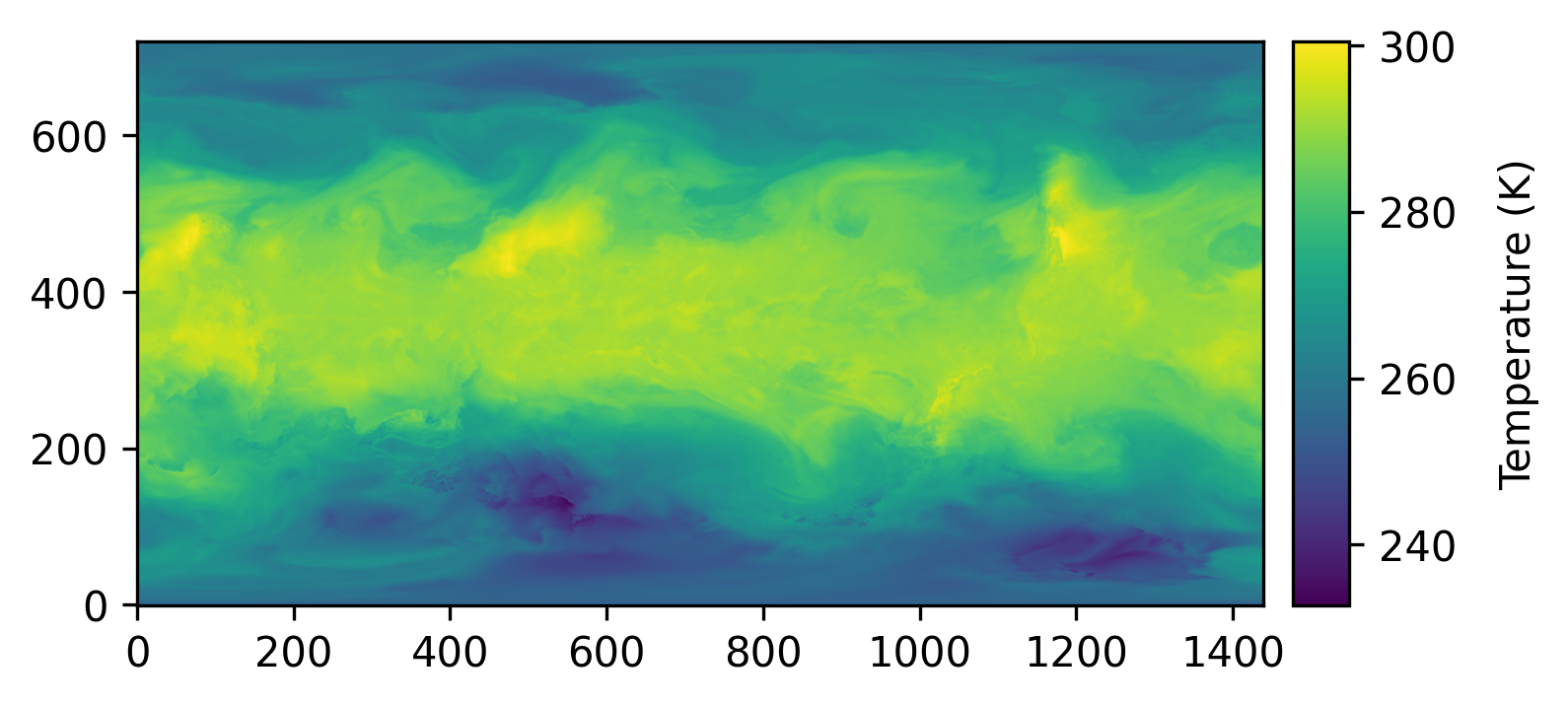}
        \caption{A single timestep}
        \label{fig:one-specific-timestamp-from-era5}
    \end{subfigure}
    \begin{subfigure}[b]{0.38\linewidth}
        \centering
        \includegraphics[width=\linewidth]{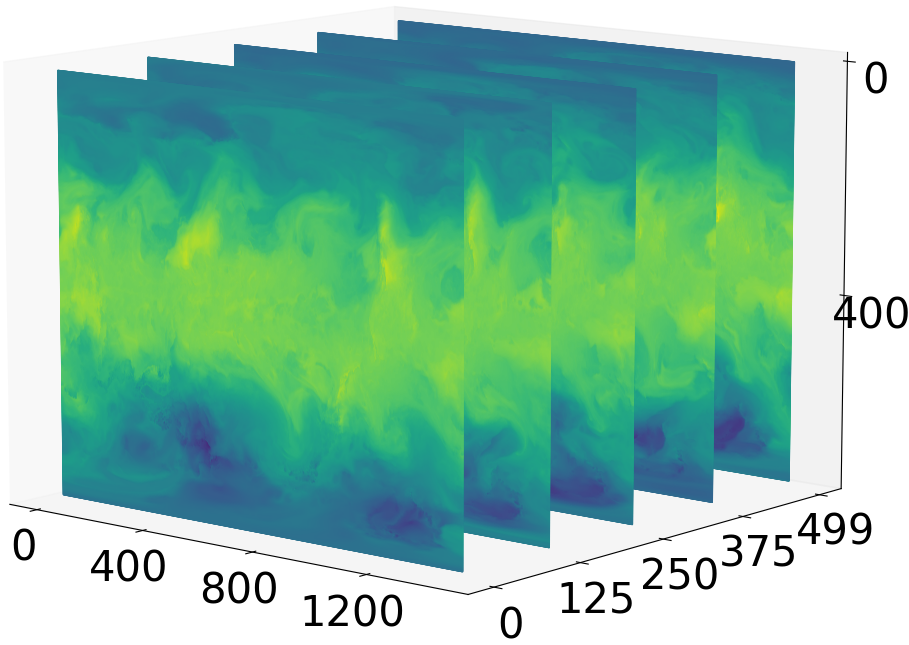} 
        \caption{500 timesteps}
        \label{fig:original-era5-ts500}        
    \end{subfigure}
    \caption{The ERA5 dataset: a visualization of ERA5 reanalysis temperature data~\cite{hersbach2020era5}}
    \label{fig:era5-temp-example}
\end{figure}

\begin{figure*}[tbp]
  \centering
  \includegraphics[width=0.9\linewidth]{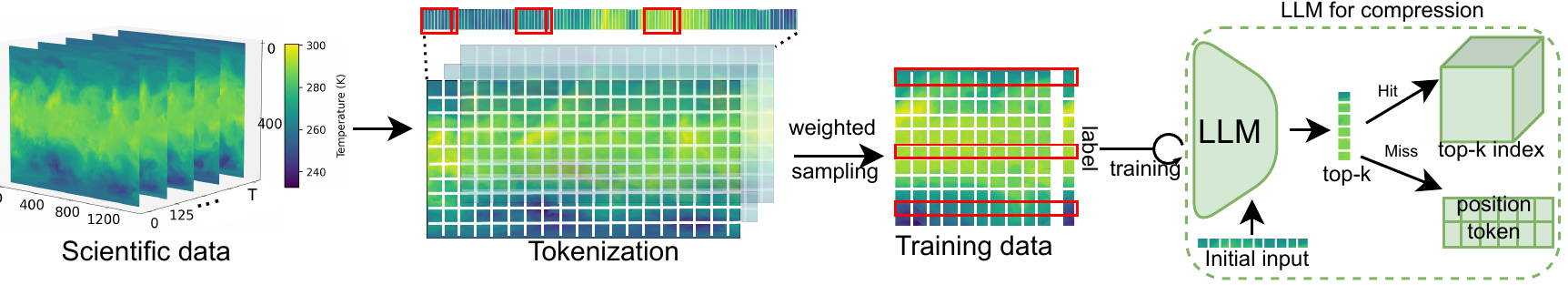}
  \caption{The workflow of \LLMComp} 
  \label{figure:workflow}
\end{figure*}

\noindent{\bf Challenges.}
Developing decoder-only LLMs for scientific data compression introduces several technical challenges.  
In this work, we focus on ERA5 climate data~\cite{hersbach2020era5} as a representative dataset shown in Figure~\ref{fig:era5-temp-example}, characterized by continuous-valued, multi-dimensional, and spatiotemporally dense fields.  
First, scientific data is inherently continuous and multi-dimensional, whereas LLMs are designed to model discrete token sequences, posing a fundamental mismatch in data representation.  
Second, large-scale spatiotemporal fields like ERA5 yield long sequences, which stress the scalability of autoregressive training and inference.  
Third, scientific compression often requires pointwise error guarantees to ensure downstream usability, fidelity constraints that are not naturally supported by standard language modeling objectives.

\noindent{\bf A new paradigm.}
In this work, we propose \LLMComp, the first framework to leverage decoder-only transformers for scientific data compression. 
We formulate this task as symbolic sequence modeling, enabling autoregressive learning of spatiotemporal structure through next-token prediction.
The workflow is illustrated in Figure~\ref{figure:workflow}.

Specifically, given a 3D scientific field (\eg a temperature tensor of shape $T \times M \times N$, Figure~\ref{fig:original-era5-ts500}), we first linearize the data into a 1D sequence by applying a space-filling curve (\eg Z-order) \emph{within each timestamp}, preserving spatial neighborhood structure while maintaining temporal ordering.
We then apply data-adaptive scalar quantization to map continuous values into token IDs with bounded quantization error. In this work, we adopt Lloyd-Max optimization~\cite{gray1998quantization} for its effectiveness on skewed distributions commonly found in scientific data, though our framework remains compatible with other quantization schemes.

The resulting token sequence is then used to train a decoder-only transformer, which autoregressively models spatiotemporal transitions with a one-step prediction objective, focusing on token dynamics rather than full-sequence language modeling.
To guarantee error-bounded reconstruction, we adopt a top-k prediction strategy during compression. If the correct token appears among the model’s top-k outputs, we store its rank index; otherwise, we store the true token and its position to ensure strict error-bounded recovery.
Finally, all stored components are further compressed using lossless coding to maximize overall compression efficiency.

We conduct comprehensive experiments on the private Redsea data~\cite{hoteit-RSRA2022} and public ERA5 data~\cite{hersbach2020era5}.
The results show that the compression ratios of our {\LLMComp} method are, in most cases, higher than the compared state-of-the-art methods (\ie  HPEZ, SZ3.1, SPERR, and ZFP), under all tested error bounds.
Compared to the second-best method, {\LLMComp} improves compression ratios up to 30\%.

\noindent{\bf Organization.}
The remainder of this paper is organized as follows.
We introduce the background in Section~\ref{sec:preliminary}.
The details of our proposed method are given in Section~\ref{sec:method}.
Section~\ref{sec:experiment} reports the experimental results.
Section~\ref{sec:Related Work} reviews the related work.
Section~\ref{sec:conclusion} concludes the paper and presents the future work.

\section{Background}\label{sec:preliminary}
This section previews the terminologies used in the paper.
Table~\ref{tab:terminologies} summarizes the main notations.

\begin{table}[t]
  \centering
  \caption{Summary of frequently used notations}
  \resizebox{.85\linewidth}{!}{
    \begin{tabular}{cl}
    \toprule
    Notation & Description \\
    \midrule
    $X$            & temporal scientific dataset $\in \mathbb{R}^{T\times M\times N}$ \\
    $T$            & number of timestamps in $X$ \\
    $M$            & number of rows in $X$ \\
    $N$            & number of columns in $X$ \\
    $\epsilon$     & user-specified relative error bound \\
    $e$            & absolute error bound, computed from $\epsilon$ \\
    $Comp, Decomp$         & compression and decompression functions \\
    $Z$            & compressed data \\
    $X'$           & decompressed data \\
    $|X|$, $|Z|$   & sizes of original and compressed data (bytes) \\
    $\rho$         & compression ratio, $\rho = |X| / |Z|$ \\
    \bottomrule
    \end{tabular} }%
  \label{tab:terminologies}%
\end{table}%

\noindent \textbf{Temporal scientific dataset $X$.} 
A temporal scientific dataset consists of $T$ timestamps, where each timestamp is represented as a matrix with $M$ rows and $N$ columns: $X \in \mathbb{R}^{T\times M\times N}$.  
Figure~\ref{fig:era5-temp-example} shows an example from the ERA5 dataset, comprising $8,760$ hourly temperature fields over a $721 \times 1,440$ global grid.

\noindent \textbf{Compression ratio $\rho$.} 
Let $|X|$ and $|Z|$ denote the sizes (in bytes) of the original and compressed data, respectively.
$|Z|$ includes all information required for decompression.
The compression ratio is formulated as:
\begin{equation}\label{eq:compression_ratio}
    \rho = \frac{|X|}{|Z|}
\end{equation}

\noindent \textbf{Problem formulation.} 
Given a temporal scientific dataset \( X \in \mathbb{R}^{T \times M \times N} \) and a user-specified relative\footnote{The absolute error bound \( e = \epsilon \cdot vrange(X) \).} error bound $\epsilon$, our goal is to design a compression-decompression pair $\langle Comp, Decomp \rangle$ that produces a compressed representation $Z = Comp(X)$ and reconstruction $X' = Decomp(Z)$, such that:

\begin{equation}\label{eq:formulation}
\begin{aligned}
    & \underset{\langle Comp, Decomp \rangle}{\arg\max} \frac{|X|}{|Z|} \\
    \text{s.t.} &\hspace{4pt} \frac{|x_i - x'_i|}{vrange(X)} \le \epsilon, \hspace{6pt} \forall x_i \in X, x'_i \in X'
\end{aligned}
\end{equation}

Here, $vrange(X)$ denotes the value range of $X$. The compression is lossy: $X’$ is an approximation of $X$ that satisfies the given error bound.

\section{LLM for Scientific Compression}\label{sec:method}
We propose a novel compression framework that reformulates scientific data compression as a token-level sequence modeling problem using large language models (LLMs). 

In particular, we first introduce the tokenization process that maps continuous values into discrete token sequences with bounded quantization error. 
We then prepare the input by linearizing 3D scientific fields into 1D sequences and incorporating spatiotemporal positional embeddings.
Next, we train a decoder-only transformer to autoregressively model token transitions under a one-step prediction objective tailored for compression.
Finally, we apply a top-$k$ prediction strategy with correction mechanisms to enable efficient and error-bounded reconstruction.

\subsection{Tokenization of Scientific Data}\label{sec:method:tokenization}
Scientific data is typically represented as high-dimensional tensors, such as 3D spatiotemporal fields of temperature in Figure~\ref{fig:original-era5-ts500}. Unlike natural language, these data are continuous-valued, spatially structured, and sampled over time, posing fundamental challenges for sequence modeling using large language models (LLMs), which require discrete token sequences as input.

To bridge this representational gap, we transform the continuous data into symbolic sequences through a two-step process: {\it spatial flattening} and {\it scalar quantization}. The goal is to generate tokenized inputs that preserve both the structural coherence of the original field and the fidelity constraints required in scientific compression.

\paragraph{Flattening via Z-order curves.}
We first linearize the 3D tensor into a 1D sequence using a spatially coherent space-filling curve, specifically, the Z-order curve. Unlike row-major or column-major flattening, Z-order traversal preserves locality by mapping spatially neighboring elements to adjacent positions in the sequence. Although any 3D-to-1D mapping inevitably introduces some distortion, we empirically show in~\cite{li2025llmcomp} that Z-order significantly improves token prediction accuracy by maintaining neighborhood coherence across space and time.

\begin{figure}[t]
    \centering
    \begin{subfigure}[b]{0.47\linewidth}
        \centering
        \includegraphics[width=\linewidth]{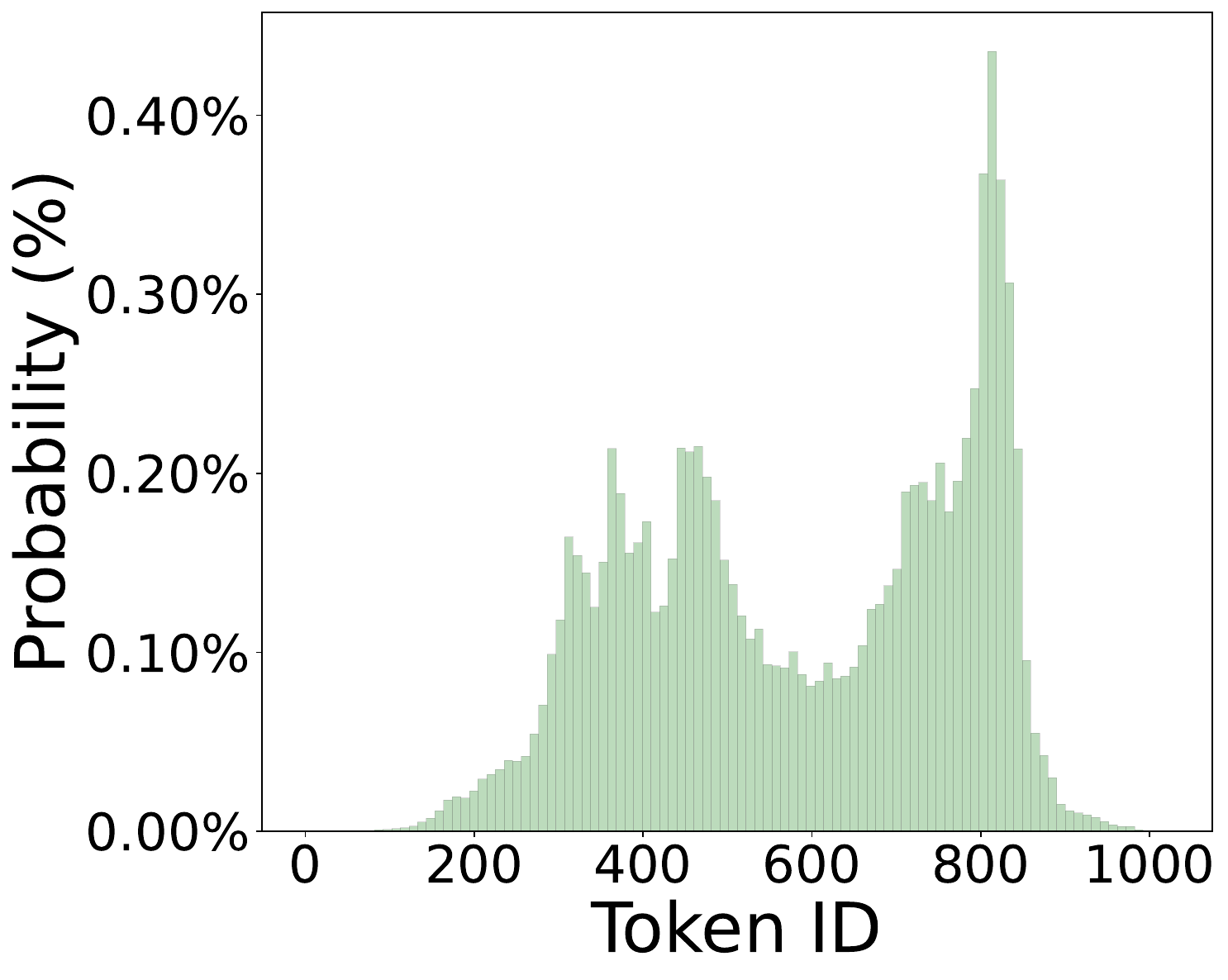}
        \caption{Original distribution}
        \label{fig:distribution-original}
    \end{subfigure}
    \begin{subfigure}[b]{0.47\linewidth}
        \centering
        \includegraphics[width=\linewidth]{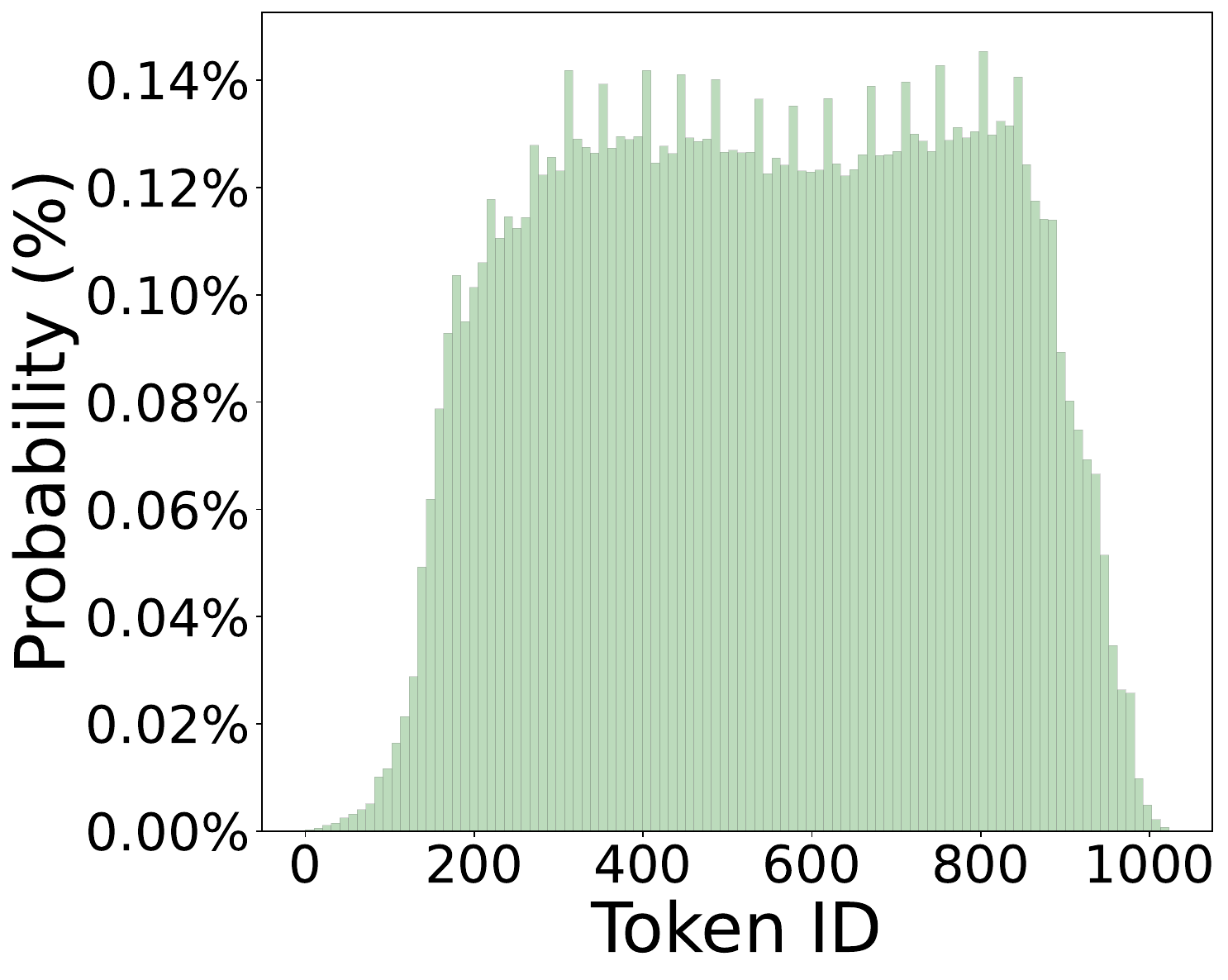}
        \caption{Target-aware sampling}
        \label{fig:distribution-target-aware}        
    \end{subfigure}
    \caption{Distribution of original token and target-aware sampling on ERA5 dataset, (a) The original token distribution is highly skewed, with dense peaks and underrepresented regions, leading to poor coverage under random or uniform sampling.
    (b) Our target-aware sampling redistributes the frequency more uniformly, ensuring rare but important tokens are adequately learned during training.}
    \label{fig:distribution-original-target-aware}
\end{figure}

\paragraph{Quantization via Lloyd-Max.}
After flattening, we apply scalar quantization to map each floating-point value to a token ID from a fixed vocabulary. We adopt the Lloyd-Max algorithm~\cite{gray1998quantization}, a classic distribution-aware method that minimizes mean reconstruction error. The algorithm begins with uniformly spaced bins and iteratively refines centroids (reconstruction levels) by computing the empirical means of assigned samples and updating bin boundaries accordingly. Compared to uniform quantization, Lloyd-Max better adapts to skewed or heavy-tailed distributions shown in Figure~\ref{fig:distribution-original}, common in scientific datasets, achieving lower distortion for a given number of levels.

This tokenization strategy is conceptually analogous to subword tokenization in NLP, converting real-valued signals into symbolic sequences. Crucially, the quantization scheme enforces a strict \emph{error bound}: each reconstructed value deviates from its original by no more than a pre-specified threshold. This property ensures that as long as the predicted token ID matches the ground truth, the final decompressed value remains within the allowed tolerance, enabling faithful, error-bounded reconstruction through autoregressive modeling.

\subsection{Spatiotemporal Positional Augmentation}\label{sec:method:input}
Flattening 3D scientific data into 1D sequences inevitably distorts the original spatial and temporal relationships, particularly at field boundaries and across time steps. While the Z-order curve partially preserves spatial locality, it cannot fully encode complex neighborhood structures inherent in spatiotemporal fields. As a result, purely sequence-based token prediction may fail to recover localized patterns that are essential for accurate compression.

To mitigate this structural distortion, we augment each token with its original spatial and temporal coordinates. 
Inspired by position encoding strategies in transformer architectures, we embed these auxiliary signals into the model’s input representation. 
Specifically, for each quantized token at position $(x, y, t)$, we define the model input vector as:
\begin{equation}
    \mathbf{h}_{x,y,t} = \mathbf{e}_{\text{token}}(v_{x,y,t}) + \mathbf{e}_{\text{space}}(x, y) + \mathbf{e}_{\text{time}}(t),
\end{equation}
where $\mathbf{e}_{\text{token}}$, $\mathbf{e}_{\text{space}}$, and $\mathbf{e}_{\text{time}}$ are learned embedding functions for the quantized value, 2D spatial coordinates, and temporal index, respectively. 
Spatial embeddings can be implemented as 2D lookup tables or learned projections of normalized coordinates.

Our structure-aware input incorporates the spatial and temporal context directly into the token representation, enabling the model to reason jointly over positional and temporal information. By aligning the token embeddings with their original $(x, y, t)$ coordinates, the model effectively captures cross-dimensional dependencies that are otherwise lost during flattening. This enriched representation significantly improves the model's ability to preserve local coherence and long-range patterns, crucial for accurate autoregressive compression in scientific domains.

\subsection{Training \LLMComp}\label{sec:method:training}
Pretrained LLMs such as GPT~\cite{achiam2023gpt}, LLaMA~\cite{touvron2023llama}, and DeepSeek~\cite{guo2025deepseek} are optimized for textual data and probabilistic generation, and are not directly applicable to scientific compression tasks that require deterministic decoding and strict fidelity guarantees. Moreover, their tokenization schemes and position encodings are incompatible with the spatiotemporal embeddings $(x, y, t)$ and quantized physical values used in our framework. 

We train a decoder-only transformer (called \LLMComp) from scratch for scientific data compression. 
The model takes as input a fixed-length context of $C$ tokens, each enriched with $(x, y, t)$ positional embeddings (Section~\ref{sec:method:input}), and is trained with a fixed-window one-step prediction strategy: it always uses the previous $C$ tokens to predict the token at position $C+1$. During training, this window slides by one position across the sequence. This design aligns explicitly with the autoregressive decoding process used in compression, reduces memory consumption, and focuses model capacity on learning token transition dynamics.

Formally, the model learns the conditional distribution:
\[
P(v_{C+1} \mid v_1, v_2, \dots, v_C),
\]
where $\{v_1, \dots, v_C\}$ is the tokenized input context extracted from the flattened sequence.

Given the large size of scientific data, it is impractical to train over all possible context windows exhaustively. Moreover, naive random sampling would cause the model to overfit to frequent tokens while underexposing rare ones shown in Figure~\ref{fig:distribution-original}, degrading compression quality. To address this, we adopt a target-aware sampling strategy: we sample fixed-length context windows such that the token at position $C{+}1$, which serves as the supervised target, is approximately uniformly distributed across the token vocabulary shown in Figure~\ref{fig:distribution-target-aware}. This ensures that both frequent and rare tokens are sufficiently represented during training, improving model generalization and stability on scientific datasets with highly imbalanced value distributions.

To jointly model token identity and physical accuracy, we adopt a hybrid loss combining classification and regression objectives. The model outputs a vocabulary-sized probability vector $\mathbf{p} \in \mathbb{R}^{V}$ for the next token. The training objective includes:

\begin{itemize}
    \item \textbf{Cross-Entropy Loss}: standard token-level supervision:
    \[
    \mathcal{L}_{\text{CE}} = -\log p_{v^\ast},
    \]
    where $v^\ast$ is the ground-truth token.
    
    \item \textbf{MSE Loss}: numeric supervision based on quantization bin midpoints. Each token $v$ corresponds to bin $[b_v, b_{v+1}]$ with midpoint $\hat{r}_v = \frac{b_v + b_{v+1}}{2}$. The MSE is:
    \[
    \mathcal{L}_{\text{MSE}} = \frac{1}{2} \left( \hat{r}_{\text{pred}} - \hat{r}_{\text{true}} \right)^2,
    \]
    where $\hat{r}_{\text{pred}}$ and $\hat{r}_{\text{true}}$ are the predicted and ground-truth midpoints.
\end{itemize}

The total loss is:
\[
\mathcal{L}_{\text{total}} = \mathcal{L}_{\text{CE}} + \alpha \cdot \mathcal{L}_{\text{MSE}},
\]
where $\alpha$ balances symbolic accuracy and physical fidelity (empirically $\alpha{=}0.1$).

Algorithm~\ref{alg:train-llm} outlines the training process of \LLMComp, where the model is optimized using a hybrid loss combining cross-entropy over token IDs and mean squared error over their corresponding physical values. 
To ensure balanced supervision across the token space, each training batch is constructed via target-aware sampling from the quantized sequences.

\begin{algorithm}[t]
\caption{Training \LLMComp}
\label{alg:train-llm}
\resizebox{\columnwidth}{!}{
\begin{minipage}{\columnwidth}
\begin{algorithmic}[1]
\REQUIRE Dataset $X \in \mathbb{R}^{T \times M \times N}$, model $\mathcal{M}_\theta$, context length $C$, learning rate $\eta$, total steps $S$
\ENSURE Trained model $\mathcal{M}_\theta$
\vspace{0.3em}
\STATE $F \leftarrow \text{Flatten}(X)$
\STATE $L \leftarrow \text{LloydMaxQuantize}(F)$ \hfill // Tokenization
\FOR{step = 1 to $S$}
    \STATE $(X_C, y) \leftarrow \text{TargetAwareSample}(L, C)$ \hfill // Context–target pairs
    \STATE $\mathbf{p} \leftarrow \mathcal{M}_\theta(X_C)$ \hfill // Vocabulary logits
    \STATE $\hat{r}_{\text{pred}} \leftarrow \text{Midpoint}(y = \arg\max \mathbf{p})$
    \STATE $\hat{r}_{\text{true}} \leftarrow \text{Midpoint}(y)$
    \STATE $\mathcal{L}_{\text{CE}} \leftarrow \text{CrossEntropy}(\mathbf{p}, y)$
    \STATE $\mathcal{L}_{\text{MSE}} \leftarrow \frac{1}{2} (\hat{r}_{\text{pred}} - \hat{r}_{\text{true}})^2$
    \STATE $\mathcal{L}_{\text{total}} \leftarrow \mathcal{L}_{\text{CE}} + \alpha \cdot \mathcal{L}_{\text{MSE}}$
    \STATE Update $\theta$ via gradient descent on $\mathcal{L}_{\text{total}}$
\ENDFOR
\RETURN $\mathcal{M}_\theta$
\end{algorithmic}
\end{minipage}
} 
\end{algorithm}

\subsection{Autoregressive Compression with top-$k$ Prediction}\label{sec:method:topk}
Once trained, \LLMComp~is used to compress scientific token sequences via autoregressive inference. Starting from an initial prefix of $C$ tokens, the model outputs a ranked top-$k$ list of candidate tokens at each decoding step. If the ground-truth token is within the top-$k$ list, we record its rank index; otherwise, we store the true token and its position in a sparse correction set.

This process ensures that compression is fully faithful: all necessary information for exact reconstruction is recorded. The top-$k$ strategy enables high compression efficiency by avoiding explicit storage of most tokens, while fallback corrections guarantee strict error-bounded recovery.

To efficiently represent prediction outcomes, we maintain a top-$k$ index array that records token ranks at each decoding step. Although compact, this index array scales linearly with sequence length. To reduce redundancy, we exploit temporal smoothness in scientific data: at fixed spatial coordinates $(x, y)$, values tend to evolve gradually over time, yielding stable token predictions. We apply delta encoding along the temporal axis, followed by lossless entropy coding, to compress the index array.

This top-$k$ guided compression ensures strict error-bounded reconstruction while avoiding the need to store every token explicitly. The index array captures predictable transitions, while the correction set handles occasional prediction errors, enabling accurate and compact representation of scientific data.

\noindent 
{\bf Decompression.} 
Given an initial prefix of $C$ tokens, decompression proceeds autoregressively using the trained \LLMComp~model. At each step, the model outputs a top-$k$ candidate list. If the corresponding top-$k$ index indicates a correct prediction (index $<k$), the corresponding token is selected directly from the model output. If the index equals $k$, the true token is retrieved from the correction set. The newly selected token is then appended to the context window to predict the next token. This iterative process continues until the entire sequence is reconstructed. Since token quantization is error-bounded by design, the resulting reconstruction satisfies the specified fidelity constraints.

\noindent
{\bf Model Storage.}
Following the practice in neural compression on scientific data~\cite{liu2021exploring,liu2023srn}, we exclude the size of the pretrained model from compression ratio calculations. The model is trained once and reused across all compression tasks without modification. In real-world deployments, it can be distributed once (\eg via offline transfer or cloud access) and amortized over long-term usage.

\begin{figure}[t]
    \centering
    \includegraphics[width=\linewidth]{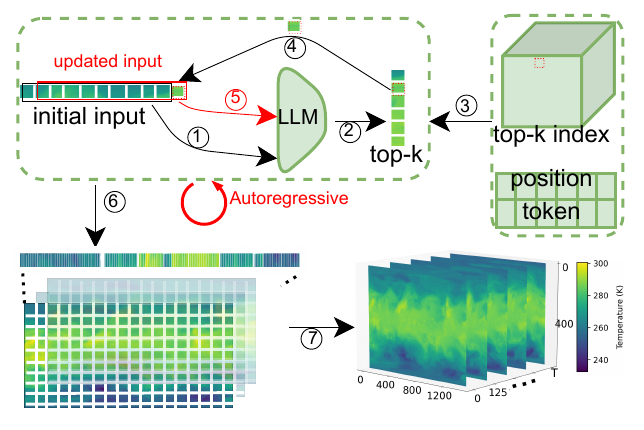}
    \caption{Autoregressive decompression workflow. Given an initial input, the LLM predicts top-$k$ tokens. The next token is reconstructed based on whether the ground truth is within top-$k$. The updated sequence is used for the next step. Recovered tokens are finally mapped to a 3D temperature volume.}
    \label{fig:reconstruction-from-llmcomp}
\end{figure}


\section{Experimental Evaluation}\label{sec:experiment}
We evaluate \LLMComp~on the Redsea and ERA5 reanalysis datasets, comparing it against four state-of-the-art (SOTA) error-bounded lossy compressors: SPERR, ZFP, SZ 3.1, and HPEZ. We report results on both compression ratio and reconstruction fidelity under varying error bounds.

\subsection{Experimental Setup}\label{sec:exp:setting}
{\bf Software and hardware.}
\LLMComp~is implemented in {\sc Python} with {\sc PyTorch}.  
Training is performed on a server with an AMD EPYC 7763 CPU (64 cores, 3.52~GHz) and four NVIDIA A100 GPU (80~GB), running Ubuntu 22.04.4 LTS (64-bit).

\stab
{\bf Datasets.}
We use two large-scale real-world data: the private Red Sea Reanalysis (RSRA)\cite{hoteit-RSRA2018, hoteit-RSRA2022} and the public ERA5\cite{hersbach2020era5}. 
These datasets cover multiple scientific variables, including temperature, humidity, wind speed, and geopotential.
Table~\ref{table:redsea-data} provides the dataset details.

\begin{table}[t]
\centering
\caption{Datasets: \myNum{i} Private: Redsea; \myNum{ii} public: ERA5; Dimensions are: $T \times M \times N$.}\label{table:redsea-data}
\resizebox{\linewidth}{!}{
\begin{tabular}{lcrcr}
\hline
 Dataset & Dimensions &  Size (GB) & Domain & Type \\
\hline
Redsea-96K & \multirow{2}{*}{96,407x855x1,215}   & \multirow{2}{*}{373.1}    & {Temperature} & {f32} \\
Redsea-96K-q2    \vspace{3pt}  &   &   &   Humidity  &  f32  \\ 
Redsea-96K-u10 & \multirow{2}{*}{96,407x855x1,215}   & \multirow{2}{*}{373.1}    & \multirow{2}{*}{Wind speed} & \multirow{2}{*}{f32} \\
Redsea-96K-v10   \vspace{3pt}   &            &            &              &    \\ 
\ERA5-2018-T & \multirow{2}{*}{8,760x721x1,440}   & \multirow{2}{*}{67.8}    & \multirow{2}{*}{Temperature} & \multirow{2}{*}{f64} \\
\ERA5-2023-T   \vspace{3pt}   &            &            &              &    \\ 
\ERA5-2018-G & \multirow{2}{*}{8,760x721x1,440}   & \multirow{2}{*}{67.8}    & \multirow{2}{*}{Geopotential} & \multirow{2}{*}{f64} \\
\ERA5-2023-G      &            &            &              &    \\ \hline
\end{tabular}}
\end{table}

\stab
{\bf Benchmarked methods.}
We compare {\LLMComp} with four state-of-the-art error-bounded lossy compressors: SPERR~\cite{li2023lossy}, ZFP~\cite{lindstrom2014fixed}, SZ 3.1~\cite{zhao2021optimizing}, and HPEZ~\cite{liu2023high}. SPERR builds on the SPECK wavelet algorithm~\cite{pearlman2004efficient,tang2006three} and supports both parallel processing and error tolerance. ZFP uses a lifted, orthogonal block transform with embedded coding to enable bit stream truncation at the block level. SZ 3.1 employs dynamic spline interpolation along with several optimizations to enhance prediction accuracy. HPEZ further advances SZ-style compression with techniques such as interpolation reordering, multidimensional interpolation, and natural cubic splines, combined with auto-tuning driven by quality metrics. We do not include AE-SZ~\cite{liu2021exploring} and SRN-SZ~\cite{liu2023srn} in our comparison, as HPEZ and SZ 3.1 already represent the strongest performing SZ-based methods to date.

\stab
{\bf Default parameters.}
We adopt a decoder-only Transformer with 12 layers, a hidden size of 768, and 12 attention heads (each of dimension 64).
Unless otherwise specified, the vocabulary size is 1024, the context length is 32, the top-$k$ value for prediction is set to $k=8$, 
and the training uses 1\% coverage-guided sampling of the full dataset.

\subsection{Overall Compression Ratio}\label{exp:results-ovreall-cr}
Table~\ref{table:final-cr-real-gan} presents the compression ratios $\rho$ across eight datasets under five relative error bounds ($\epsilon = 10^{-2}$ to $10^{-6}$). 
Our method consistently achieves the highest compression ratio across all datasets and error bounds (except under $10^{-2}$), demonstrating its robustness and superior adaptability.

At loose error bounds (\eg $\epsilon = 10^{-2}$), traditional compressors such as SZ3.1 and HPEZ perform competitively, occasionally achieving slightly higher ratios due to their highly tuned prediction mechanisms.
For instance, SZ3.1 reaches $860.5\times$ on \ERA5-2023-G, outperforming our method in this specific case.
However, these methods experience a sharp performance drop as the error bound tightens.

In contrast, our method maintains high performance, especially under practically relevant regimes like $\epsilon = 10^{-3}$ to $10^{-5}$. 
For example, on \ERA5-2023-T at $\epsilon = 10^{-3}$, our method achieves a ratio of $37.80\times$, outperforming the best baseline HPEZ ($32.59\times$) by $16\%$.
Even under stringent constraints ($\epsilon = 10^{-6}$), our method retains its edge, consistently surpassing all competitors across datasets.

In summary, our method provides state-of-the-art compression performance across a wide range of scientific datasets, adapting well to diverse variable types, spatial resolutions, and fidelity constraints. These results confirm its practical effectiveness and generalizability in real-world scientific data compression tasks.

\begin{table}[t]
\caption{Compression ratio $\rho$ for various datasets and error bounds. \textbf{Bold} values indicate best performance; \underline{underlined} values are runner-ups.}
\label{table:final-cr-real-gan}
\resizebox{\linewidth}{!}{
\begin{tabular}{lcrrrrr}
\hline
\bf  Dataset & $\epsilon$ & \bf SPERR & \bf ZFP  & \bf SZ3.1 & \bf HPEZ  & Ours \\
\hline\hline
        & $10^{-2}$ & 92.54 & 8.53 & {\bf 105.21} & \underline{102.59} &  57.13  \\
        & $10^{-3}$ & 14.20 & 4.15 & 14.94 & \underline{15.29} &  {\bf 16.10}   \\
Redsea-96K  & $10^{-4}$ & 5.85  & 2.99 & \underline{6.13} & 5.95  &  {\bf 6.53}    \\
        & $10^{-5}$ & 3.20  & 2.12 & 3.31   & \underline{3.42}  &  {\bf 3.72}    \\ 
        & $10^{-6}$ & 2.96 & 2.01 & 3.02  & \underline{3.05} &  {\bf 3.24}    \\ \hline
        & $10^{-2}$ & 61.86 & 5.97 & \textbf{79.63}  & \underline{74.06} &  42.05    \\
        & $10^{-3}$ & 11.28 & 3.83 & 12.70 & \underline{12.77} &  {\bf 14.12}   \\
Redsea-96K-q2 & $10^{-4}$ & 5.28  & 2.59 & 5.54  & \underline{5.64}  &  {\bf 6.04}    \\
        & $10^{-5}$ & 3.00  & 1.98 & \underline{3.25}  & 3.11  &  {\bf 3.65}    \\ 
        & $10^{-6}$ & 2.79 & 1.90 & \underline{2.98}  & 2.76 &  {\bf 3.16}    \\ \hline
        & $10^{-2}$  & 63.39 & 6.70 & \textbf{65.70}  & \underline{63.99} &  39.94  \\
        & $10^{-3}$ & 11.35 & 4.12 & 11.44 & \underline{12.00} &  {\bf 13.12}   \\
Redsea-96K-u10 & $10^{-4}$ & 5.07  & 2.72 & 5.22  & \underline{5.45}  &  {\bf 5.90}    \\
        & $10^{-5}$ & 2.91  & 2.01 & 2.99  & \underline{3.08}  &  {\bf 3.41}    \\ 
        & $10^{-6}$ & 2.68 & 1.84 & \underline{2.78}  & 2.72 &  {\bf 3.17}    \\ \hline
        & $10^{-2}$  & 55.77 & 6.63 & \textbf{79.24}  & \underline{74.41} &  43.52   \\
        & $10^{-3}$ & 10.68 & 4.09 & 12.48 & \underline{13.10} &  {\bf 14.35}   \\
Redsea-96K-v10 & $10^{-4}$ & 5.22  & 2.92 & 5.45  & \underline{5.63}  &  {\bf 6.16}    \\
        & $10^{-5}$ & 2.95  & 2.07 & 3.10  & \underline{3.21}  &  {\bf 4.07}    \\ 
        & $10^{-6}$ & 2.72 & 1.95 & \underline{2.93}  & 2.89 &  {\bf 3.69}    \\ \hline
         & $10^{-2}$  & 219.81& 9.98 & \textbf{370.19} & \underline{293.15}  &  {75.27}  \\
        & $10^{-3}$ & 30.16 & 5.41 & 32.29 & \underline{33.39} &  {\bf 38.13}   \\
\ERA5-2018-T & $10^{-4}$ & 9.52  & 3.59 & 9.18  & \underline{9.71}  &  {\bf 11.06}   \\
        & $10^{-5}$ & 4.51  & 2.59 & 4.60  & \underline{4.98}  &  {\bf 5.28}    \\ 
        & $10^{-6}$ & 4.21 & 2.23 & 4.18  & \underline{4.47} &  {\bf 4.82}    \\ \hline
         & $10^{-2}$  & 221.15& 9.95 & \textbf{374.94} & \underline{295.06} &  80.09  \\
        & $10^{-3}$ & 30.31 & 5.39 & 32.46 & \underline{32.59} &  {\bf 37.80}   \\
\ERA5-2023-T & $10^{-4}$ & 9.55  & 3.59 & 9.24  & \underline{9.56}  &  {\bf 10.71}   \\
        & $10^{-5}$ & 4.58  & 2.61 & 4.63  & \underline{4.85}  &  {\bf 4.97}    \\ 
        & $10^{-6}$ & 4.24 & 2.34 & 4.25  & \underline{4.33} &  {\bf 4.49}    \\ \hline
        & $10^{-2}$  & 569.38& 13.19&\underline{858.46} & 849.84 &  {283.66} \\
        & $10^{-3}$ & 56.30  & 7.09 & 79.95 & \underline{80.01} &  {\bf 86.17}  \\
\ERA5-2018-G & $10^{-4}$ & 17.28 & 4.29 & 15.91 & \underline{18.49} &  {\bf 20.03}   \\
        & $10^{-5}$ & 8.53  & 3.21 & 7.61  & \underline{8.89}  &  {\bf 9.69}   \\ 
        & $10^{-6}$ & 6.79 & 2.99 & 6.13  & \underline{7.01} &  {\bf 7.45}    \\ \hline
        & $10^{-2}$  & 572.89& 13.8 & \textbf{860.51} & \underline{852.26} & 291.92  \\
        & $10^{-3}$ & 55.19 & 7.29 & \underline{80.23} & 80.12 & {\bf 85.03}   \\
\ERA5-2023-G & $10^{-4}$ & 17.71 & 4.38 & 15.92 & \underline{19.65} & {\bf 20.93}    \\
        & $10^{-5}$ & 8.74  & 3.35 & 7.69  & \underline{9.11}  & {\bf 9.64}    \\ 
        & $10^{-6}$ & 6.93 & 3.02 & 6.45  & \underline{8.25} &  {\bf 8.96}    \\ \hline
\hline
\end{tabular}}
\end{table}

\subsection{Rate distortion analysis}

\begin{figure}[t]
    \centering
    \begin{subfigure}[b]{.9\linewidth}
        \centering
        \includegraphics[width=\linewidth]{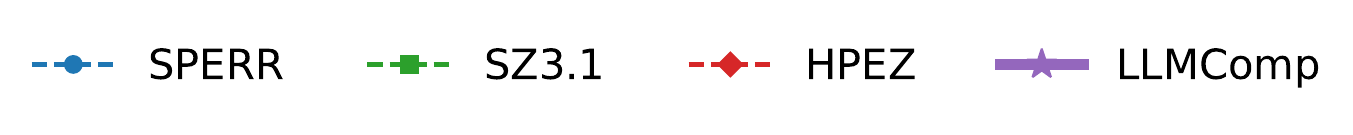}
    \end{subfigure}
    \begin{subfigure}[b]{0.49\linewidth}
        \centering
        \includegraphics[width=\linewidth]{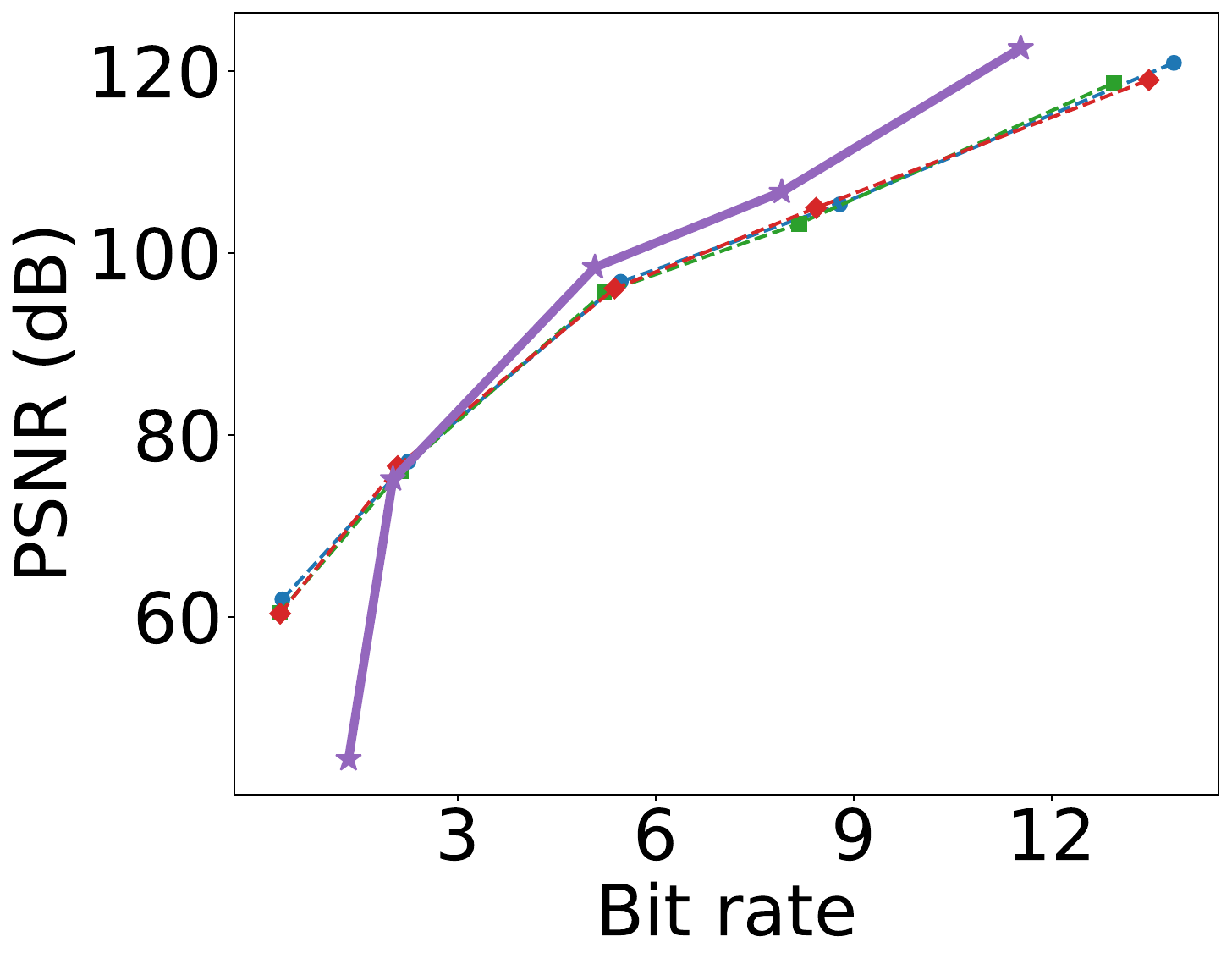}
        \caption{\RS96K}
    \end{subfigure}%
    \begin{subfigure}[b]{0.49\linewidth}
        \centering
        \includegraphics[width=\linewidth]{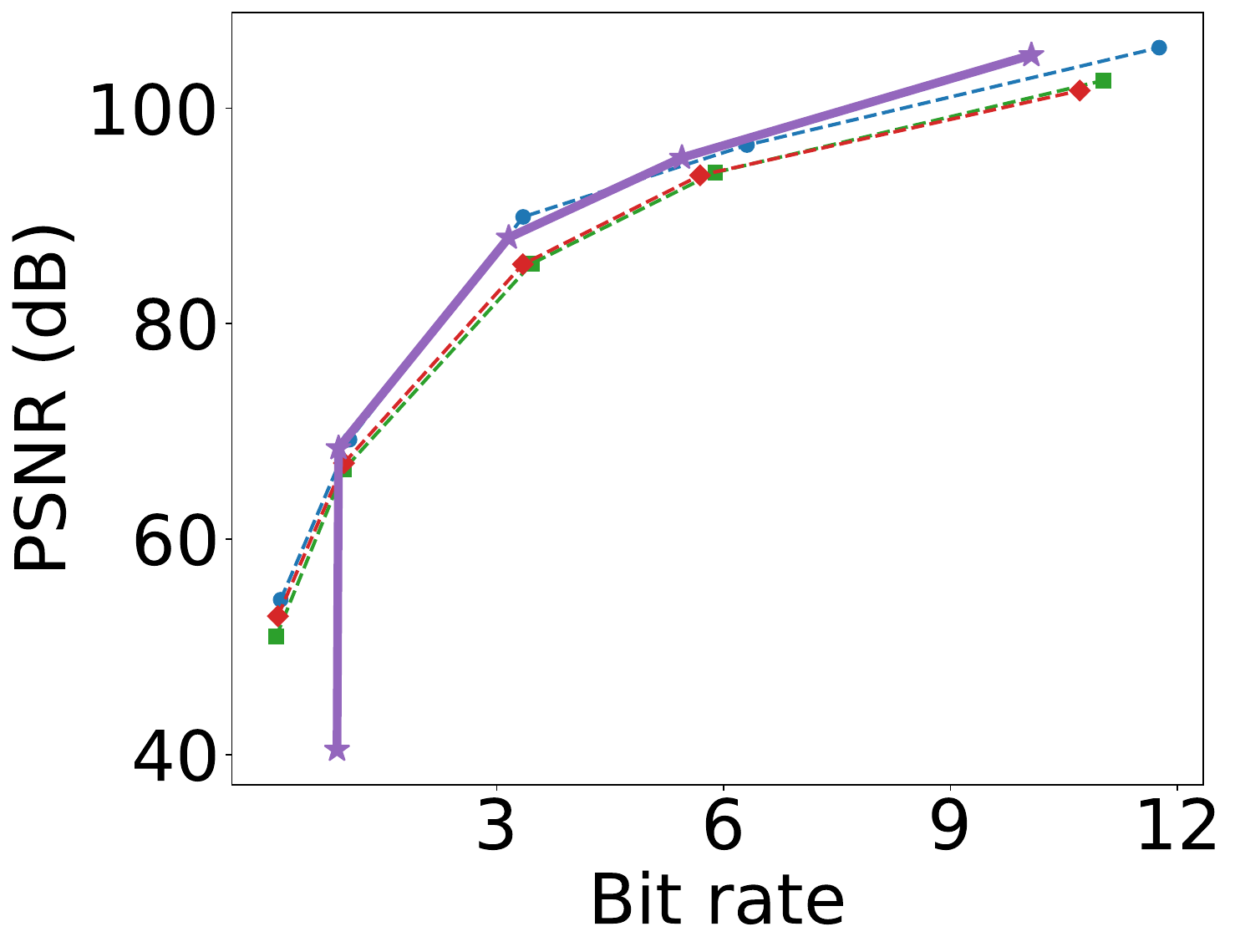}
        \caption{\ERA5-23-T}
    \end{subfigure}%
    \caption{Decompression quality (PSNR -- larger is better) vs. efficiency (bit rate -- lower is better).}
     \label{fig:exp:psnr-analysis}
\end{figure}

\noindent
We evaluate decompression fidelity using peak signal-to-noise ratio (PSNR; higher is better) and compression efficiency using bit rate (lower is better).
Figure~\ref{fig:exp:psnr-analysis}\footnote{We omit ZFP curves for clarity, as their significantly higher bitrates compress the dynamic range of the figure and obscure trends among stronger baselines.} illustrates the rate–distortion trade-offs across five compressors on the \RS96K and \ERA5-2023-T datasets.

Across both datasets, \LLMComp~consistently achieves favorable PSNR at substantially lower bitrates. 
On \RS96K, \LLMComp~achieves 122.54~dB at 11.53~bpp, surpassing SZ3.1 (118.73~dB at 12.94~bpp) and HPEZ (119.04~dB at 13.47~bpp), offering better reconstruction quality and lower storage cost. 
Similarly, on \ERA5-2023-T, \LLMComp~achieves 104.93~dB at 10.07~bpp, outperforming SZ3.1 (102.57~dB at 11.02~bpp) and HPEZ (101.64~dB at 10.71~bpp).
At looser error bounds, the advantage becomes more pronounced. 
For example, under high error tolerance, \LLMComp~achieves 40.46~dB at 0.89~bpp, while SZ3.1 requires over 1.5$\times$ the bitrate to reach similar PSNR. 
These gains persist across all distortion regimes.

Overall, these results demonstrate that \LLMComp~offers state-of-the-art rate–distortion performance, balancing fidelity and storage efficiency across a wide range of scientific data.

\begin{figure}[t]
    \centering
    \begin{subfigure}[b]{0.7\linewidth}
        \centering
        \includegraphics[width=\linewidth]{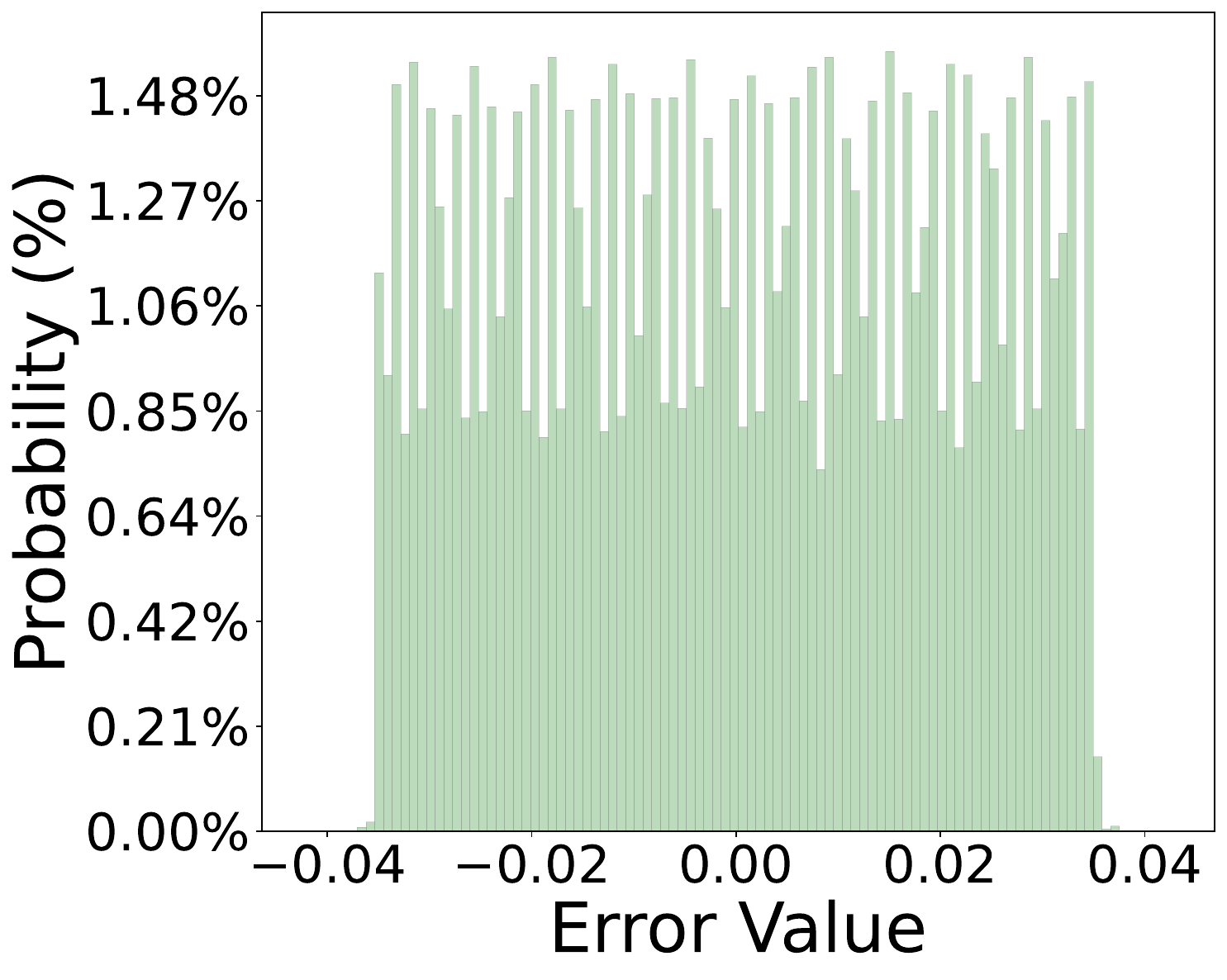}
    \end{subfigure}%
    \caption{Absolute difference between original and decompressed data for the \ERA5-23-T dataset on vocabulary size 1024.}
    \label{fig:exp:vocabulary_size}
\end{figure}

\stab
{\bf Error analysis.}
To understand the behaviour of our method, we analyse the decompression error. 
Figure~\ref{fig:exp:vocabulary_size} shows the distribution of the point-wise absolute decompression error under a vocabulary size of 1024. 
The error distribution indicates that our quantized token-based representation yields highly accurate reconstruction, with nearly all point-wise errors confined within the range of $[-0.04, 0.04]$.
Intuitively, we aim for most errors to be both small in magnitude and highly concentrated around zero, resulting in a sharp central peak in the distribution. 
This confirms the effectiveness of our approach in preserving high-fidelity temperature values even under aggressive compression.

\subsection{Effect of Top-k}\label{sec:exp:topk}
\begin{figure}[t]
    \centering
    \begin{subfigure}[b]{.5\linewidth}
        \centering
        \includegraphics[width=\linewidth]{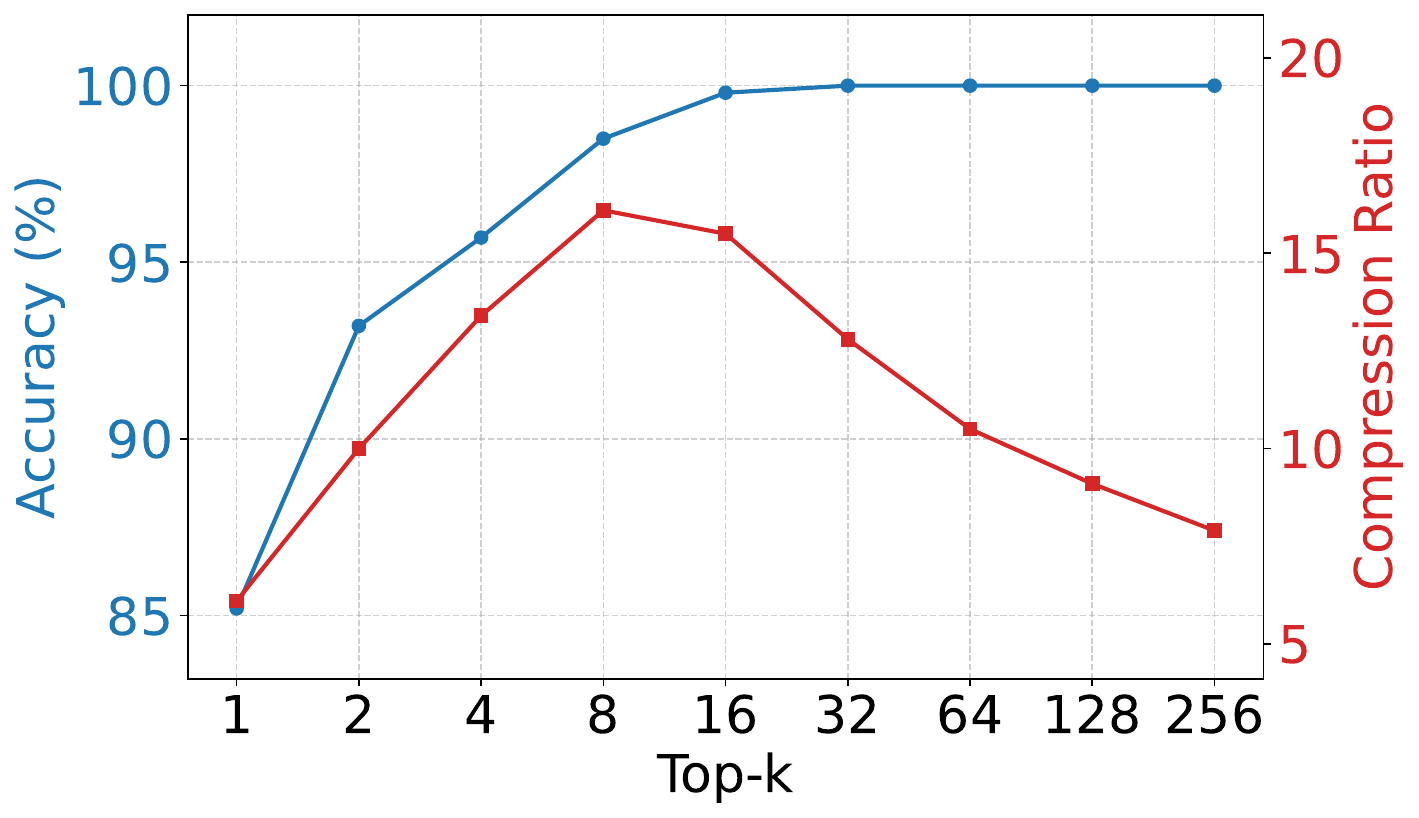}
        \caption{Redsea-96K}\label{fig:exp:topk-redsea96k}
    \end{subfigure}%
    \begin{subfigure}[b]{.5\linewidth}
        \centering
        \includegraphics[width=\linewidth]{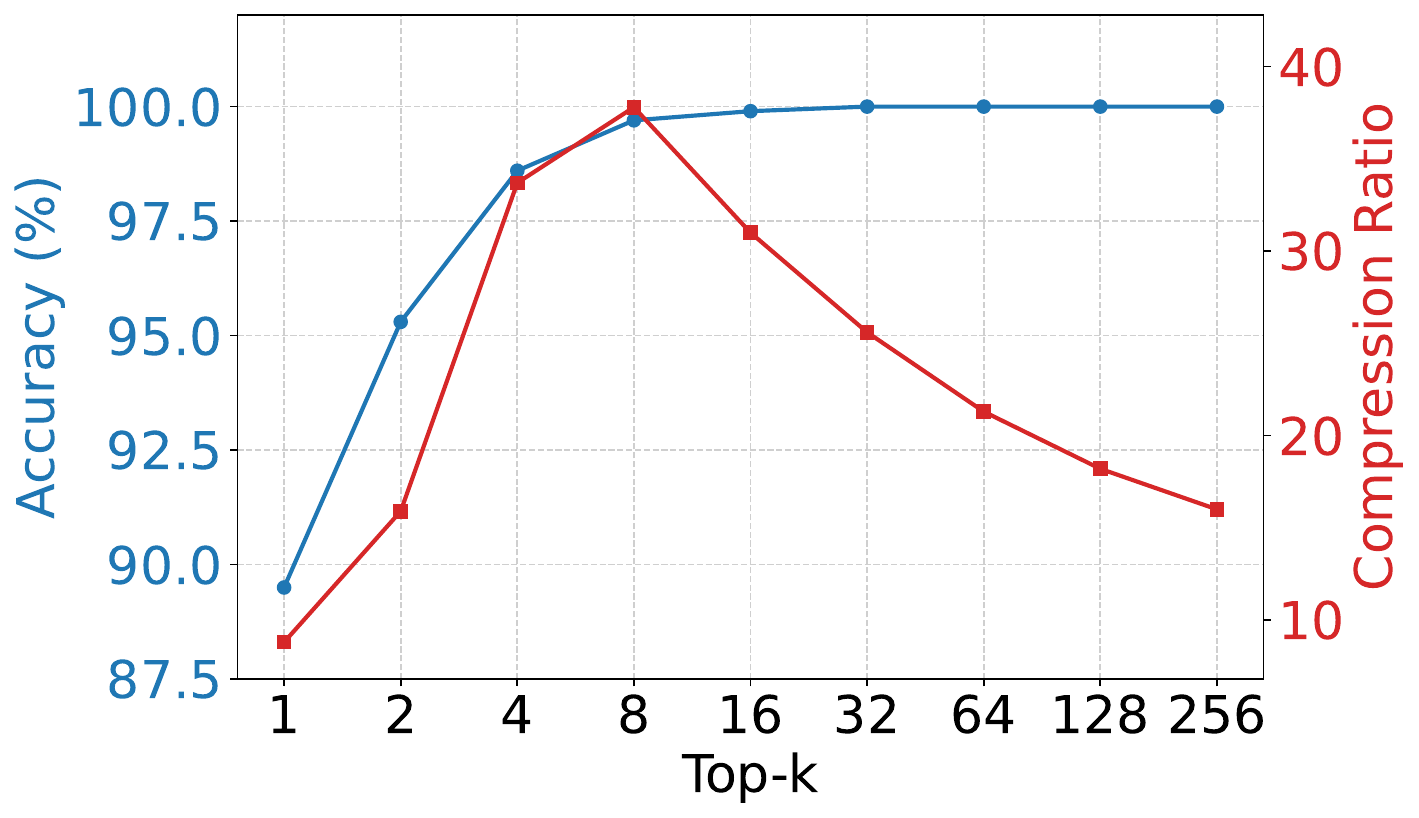}
        \caption{\ERA5-23-T}\label{fig:exp:topk-era5-2023-t}
    \end{subfigure}%
    \caption{Accuracy and Compression Ratio vs. Top-k ($\epsilon = 10^{-3}$).}
\end{figure}

We evaluate the impact of Top-k on both accuracy and compression ratio $\rho$. 
Figure~\ref{fig:exp:topk-redsea96k} (Redsea-96K) and Figure~\ref{fig:exp:topk-era5-2023-t} (\ERA-2023-T) 
present the trends of accuracy (left axis) and compression ratio (right axis) as k increases.

For both datasets, accuracy improves rapidly and reaches saturation at relatively small values of top-$k$. 
Redsea-96K achieves nearly top accuracy by $k{=}16$, and reaches full accuracy by $k{=}32$. 
\ERA5-2023-T reaches full accuracy by $k{=}32$ as well.
In contrast, the compression ratio exhibits a peak at moderate k values and then gradually decreases. 
On Redsea-96K, the best compression occurs around $k{=}8$, yielding a ratio of approximately 16.1. 
ERA5-2023-T peaks at $k{=}8$, with a maximum compression ratio of 37.8, followed by a gradual decline.

These results highlight a trade-off between accuracy and compression. 
Larger top-$k$ improves prediction accuracy and reduces fallback storage, but also increases the entropy of the index array. 
Beyond a certain point, higher top-$k$ yields diminishing returns in compression ratio despite perfect accuracy.

\begin{table*}[ht]
\centering
\caption{Impact of vocabulary size on accuracy and storage components for \ERA5-2023-T ($\epsilon = 10^{-3}$)}
\label{tab:vocab-size-breakdown}
\resizebox{\linewidth}{!}{
\begin{tabular}{cc|cccc|c}
\toprule
\textbf{Vocabulary size} & \textbf{Acc (\%)} 
& \textbf{Top-k storage (k = 8)} & \textbf{Real token storage} & \textbf{Error-Bound storage} & \textbf{All storage} 
& \textbf{Compression ratio} \\
& & \multicolumn{4}{c|}{(GB)} & \\
\midrule
128  & 100.0 & 1.59 & 0.000  & 2.784  & 4.37  & 15.5 \\
256  & 100.0 & 1.59 & 0.000  & 1.454  & 3.04  & 22.2 \\
512  & 99.9  & 1.59 & 0.067  & 0.875  & 2.53  & 26.7 \\
1024 & 99.7  & 1.59 & 0.203  & 0.000  & 1.79  & 37.8 \\
2048 & 95.7  & 1.59 & 2.915  & 0.000  & 4.50  & 15.0 \\
\bottomrule
\end{tabular}}
\end{table*}

\subsection{Effect of Vocabulary Size}\label{sec:exp:vocabularysize}
We analyze the effect of vocabulary size on compression performance using \ERA5-2023-T under the error bound $\epsilon = 10^{-3}$. 
Table~\ref{tab:vocab-size-breakdown} presents the accuracy, individual storage components, and overall compression ratio for varying vocabulary sizes.

As the vocabulary size increases from 128 to 2048, we observe two opposing effects. On one hand, a larger vocabulary provides finer quantization, reducing the error-bound storage significantly, from 2.784GB (128) to zero (1024+). On the other hand, prediction accuracy gradually drops, from 100.0\% at 128–256 to 95.7\% at 2048, leading to increased fallback storage. This trade-off causes the fallback cost (real token storage) to grow sharply beyond 1024, offsetting the benefits of finer resolution.

The optimal point occurs at vocabulary size 1024, where a near-perfect accuracy (99.7\%) is retained, and error-bound storage drops to zero. This balance results in the lowest overall storage (1.79GB) and the highest compression ratio of 37.8$\times$.
Beyond this point, increasing vocabulary further to 2048 causes a significant rise in fallback cost (2.915GB), lowering the compression ratio back to 15.0$\times$, comparable to the smallest vocabulary setting.

These results highlight a key insight: an appropriate vocabulary size is essential for balancing quantization resolution and model prediction accuracy. Overly large vocabularies may harm compressibility due to increased model uncertainty and decoding overhead.

\subsection{Effect of Context Length}
We study how the context length, namely the number of preceding tokens used for prediction, affects model performance and efficiency on the \ERA5-2023-T dataset.

\begin{figure}[t]
    \centering
    \begin{subfigure}[b]{0.7\linewidth}
        \centering
        \includegraphics[width=\linewidth]{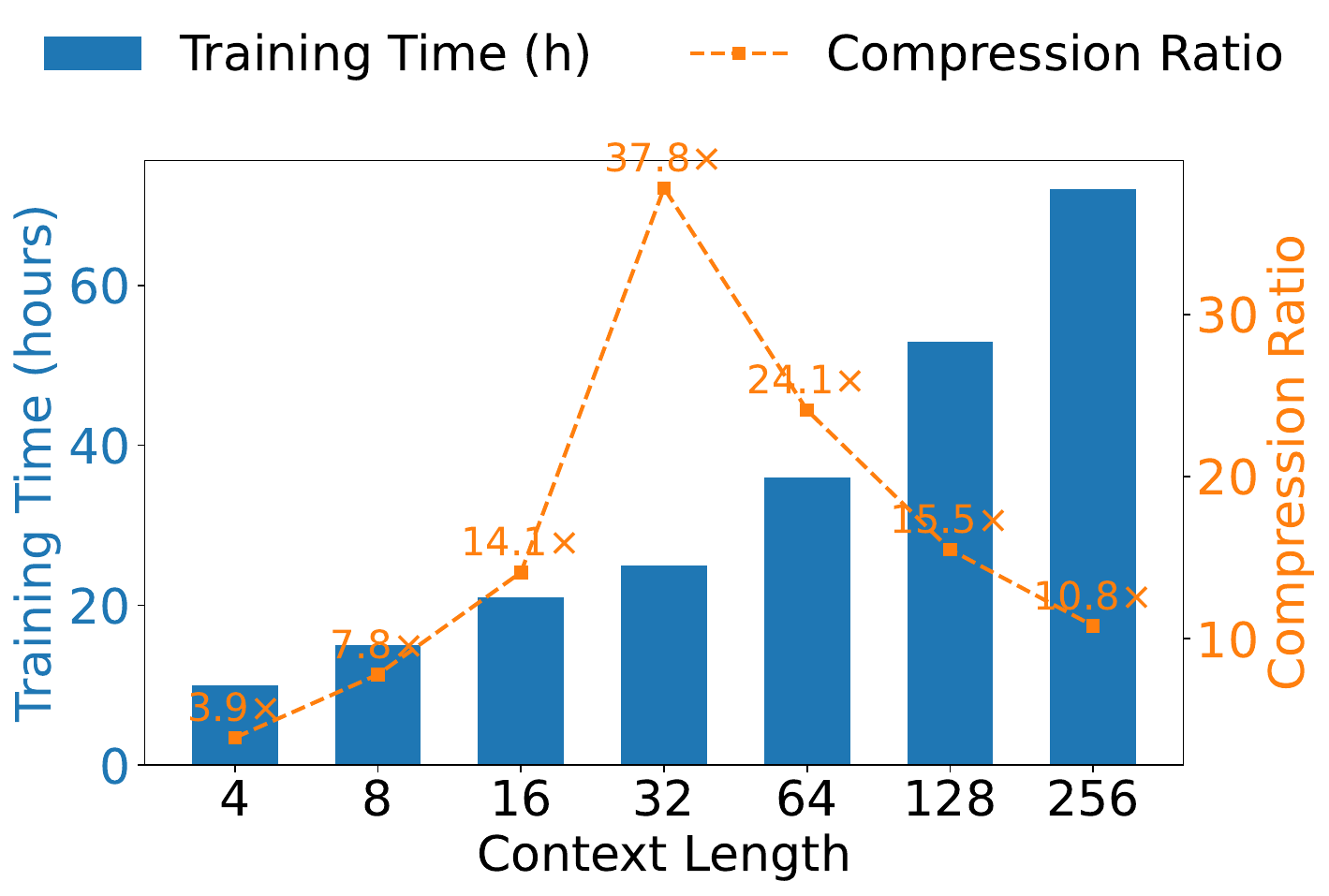}
    \end{subfigure}%
    \caption{Effect of context length on accuracy, training time, and compression ratio. Accuracy and compression improve rapidly up to length 32, beyond which gains saturate while training cost increases substantially}
    \label{fig:exp:context-length}
\end{figure}

As shown in Figure~\ref{fig:exp:context-length}, increasing the context length from 4 to 32 yields substantial gains in accuracy (from 77.1\% to 99.7\%) and compression ratio (from $3.9\times$ to $37.8\times$), mainly due to improved prediction and reduced fallback storage. However, further increasing the length beyond 32 leads to diminishing returns: accuracy and compression ratio drop, while training time rises sharply, reaching 72 hours at length 256.
These results suggest that most temporal dependencies lie in the moderate range, and longer contexts may introduce noise or overfitting. 
A moderate length (\eg 32) achieves the best trade-off between compression ratio and training cost.

\subsection{Effect of Sampling Strategies}\label{sec:exp:sampling}
\begin{figure}[t]
    \centering
    \begin{subfigure}[b]{0.9\linewidth}
        \centering
        \includegraphics[width=\linewidth]{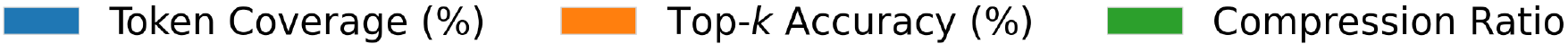}
    \end{subfigure}

    \vspace{0.5em}

    \begin{subfigure}[b]{0.48\linewidth}
        \centering
        \includegraphics[width=\linewidth]{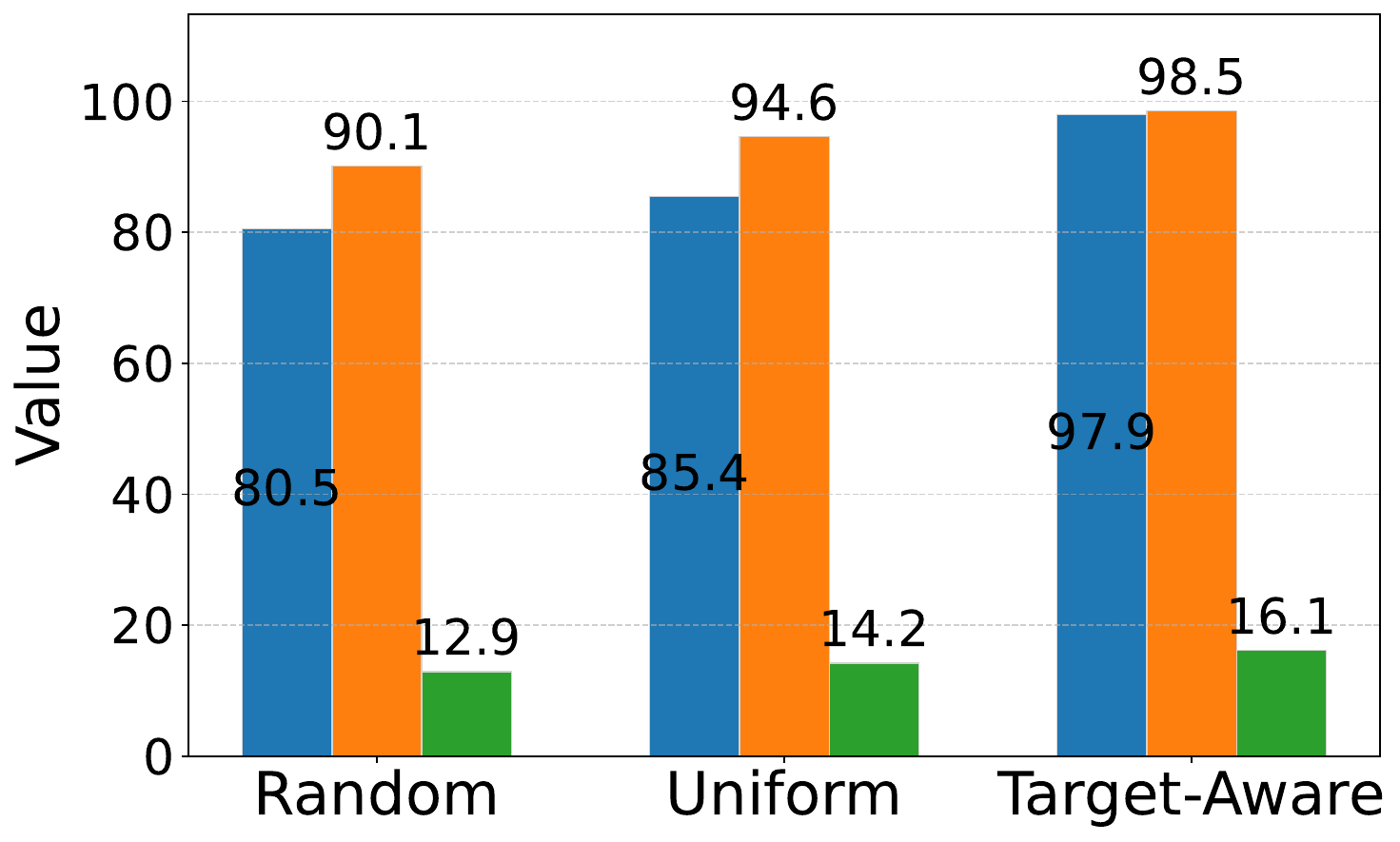}
        \caption{Redsea-96K}\label{fig:exp:sampling-redsea96k}
    \end{subfigure}
    \hfill
    \begin{subfigure}[b]{0.48\linewidth}
        \centering
        \includegraphics[width=\linewidth]{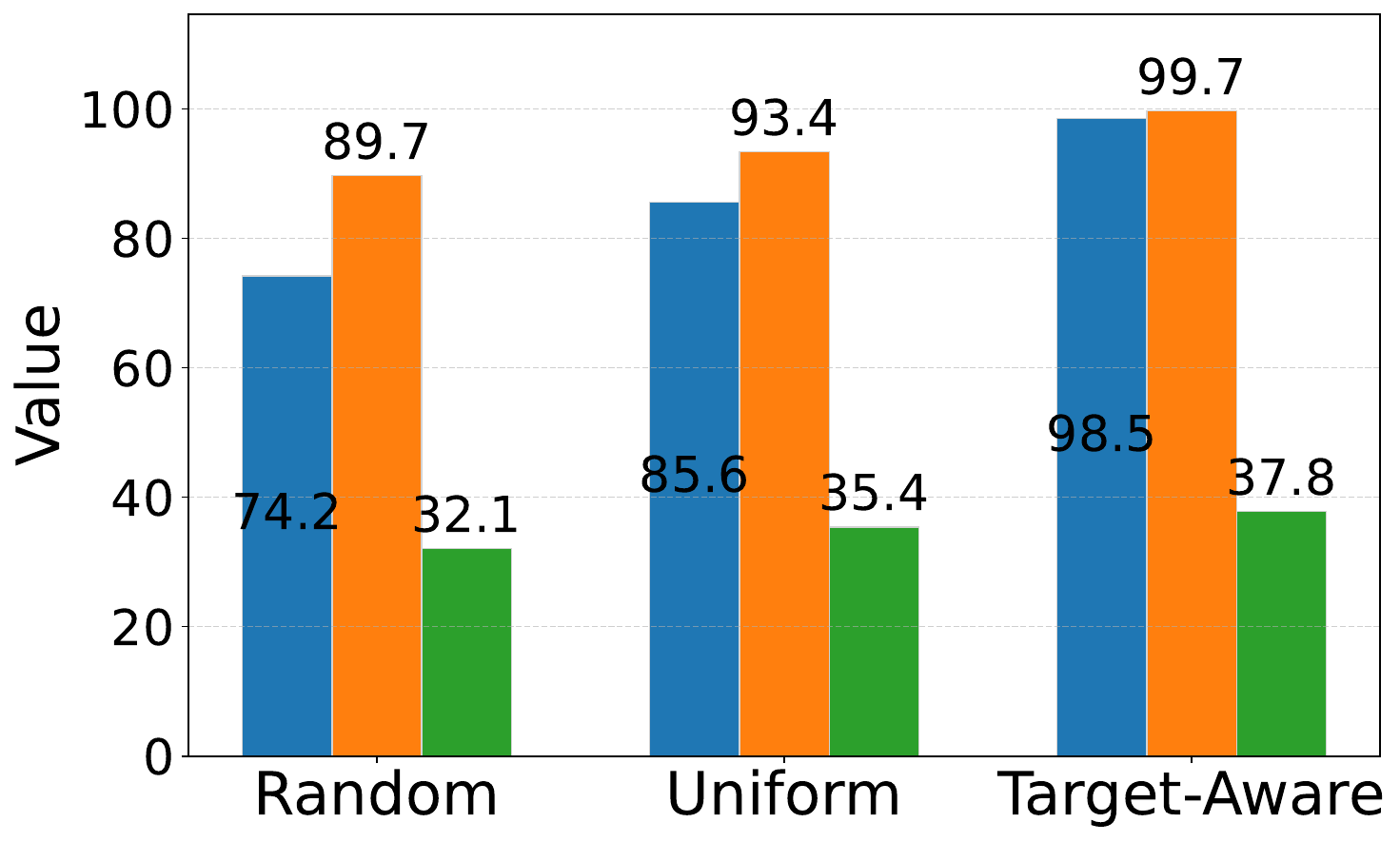}
        \caption{\ERA5-23-T}\label{fig:exp:sampling-era5-2023-t}
    \end{subfigure}

    \caption{Comparison of sampling strategies on compression performance ($\epsilon = 10^{-3}$).}
    \label{fig:exp:sampling}
\end{figure}

We evaluate how different sampling strategies impact compression under fixed vocabulary size 1024 and top-\(k\) decoding (Redsea-96K: \(k=8\), \ERA5-23-T: \(k=8\)). 
Figure~\ref{fig:exp:sampling} compares random, uniform, and target-aware sampling across three metrics: token coverage, top-\(k\) accuracy, and compression ratio.

Target-aware sampling consistently achieves the best results across both datasets. 
On \ERA5-23-T, it reaches 98.5\% token coverage, 99.7\% top-\(k\) accuracy, and a compression ratio of 37.8, compared to 74.2\% coverage and 32.1 compression ratio with random sampling. 
On Redsea-96K, it yields 97.9\% coverage and a compression ratio of 16.1, outperforming random (12.9) and uniform (14.2).

Target-aware sampling prioritizes training examples that expose the model to diverse and frequently mispredicted tokens, thereby improving prediction robustness. 
These results highlight the importance of training token diversity: higher coverage improves prediction accuracy and reduces fallback storage (namely storage for real tokens when predictions fail), enabling more efficient compression. 
In practice, even simple heuristics like target-aware sampling can significantly enhance model generalization and mitigate systematic errors on hard tokens.

\subsection{Effect of Sampling Ratio}
We investigate how the proportion of training data impacts model accuracy, compression ratio, and training cost. 

\begin{figure}[t]
    \centering
    \includegraphics[width=.8\linewidth]{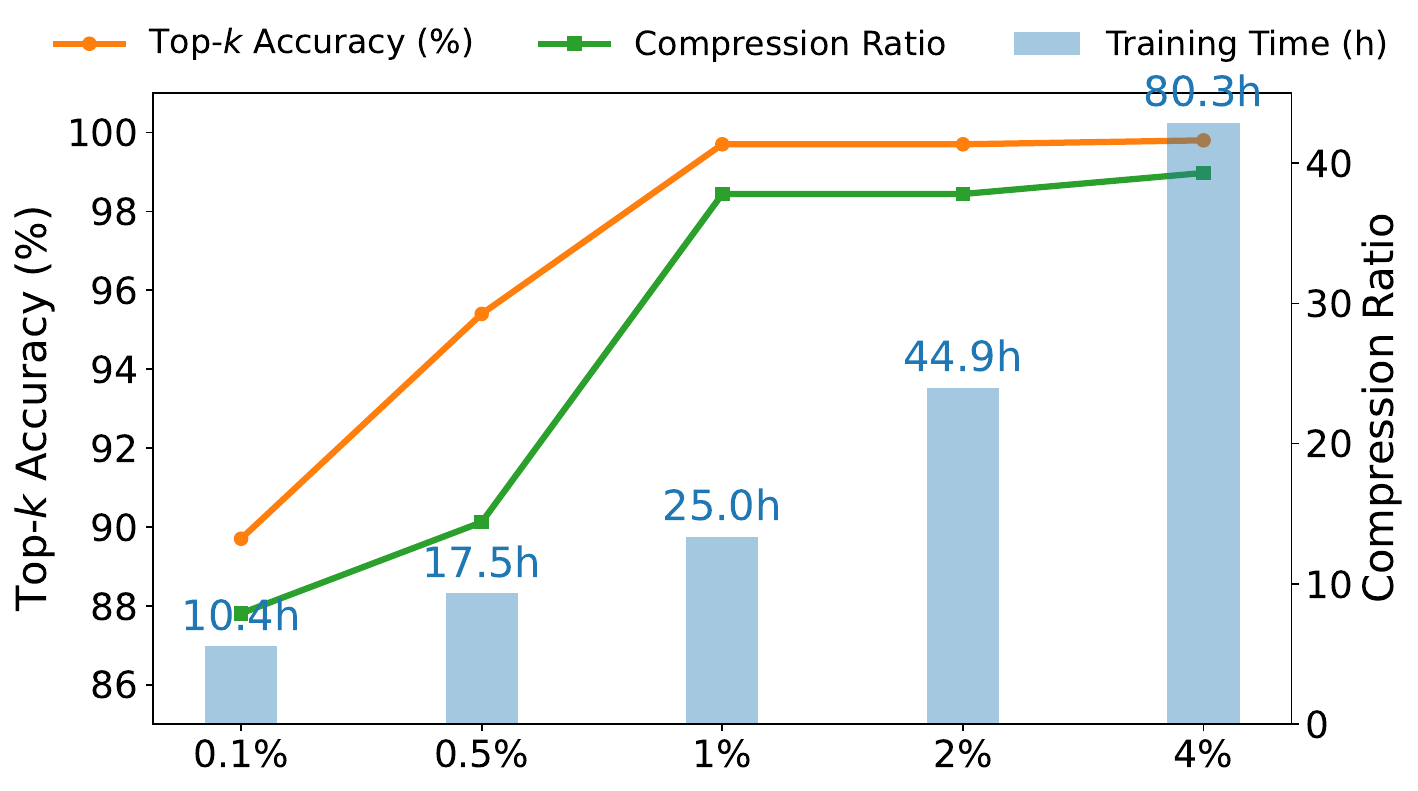}
    \caption{Effect of sampling ratio on accuracy, storage, and compression ratio on \ERA5-23-T ($k=8$, vocabulary size=1024, $\epsilon = 10^{-3}$).}
    \label{fig:exp:sampling-ratio}
\end{figure}

As shown in Figure~\ref{fig:exp:sampling-ratio}, increasing the sampling ratio from $0.1\%$ to $1\%$ significantly improves both accuracy (from $89.7\%$ to $99.7\%$) and compression ratio (from $7.9\times$ to $37.8\times$). This is attributed to the model gaining better generalization ability and reducing fallback storage as more representative patterns are observed.
Beyond $1\%$, both metrics begin to saturate: accuracy improves marginally to $99.8\%$ at $4\%$, and compression ratio grows slightly to $39.3\times$. However, the training time rises substantially, from 25 hours at $1\%$ to 80 hours at $4\%$, reflecting the increased computational burden.

Overall, the results suggest that a sampling ratio around $1\%$ offers a strong balance between training cost and compression quality. Larger samples yield diminishing returns, highlighting the data efficiency of our method.

\begin{figure}[t]
    \centering
    \begin{subfigure}[b]{0.8\linewidth}
        \centering
        \includegraphics[width=\linewidth]{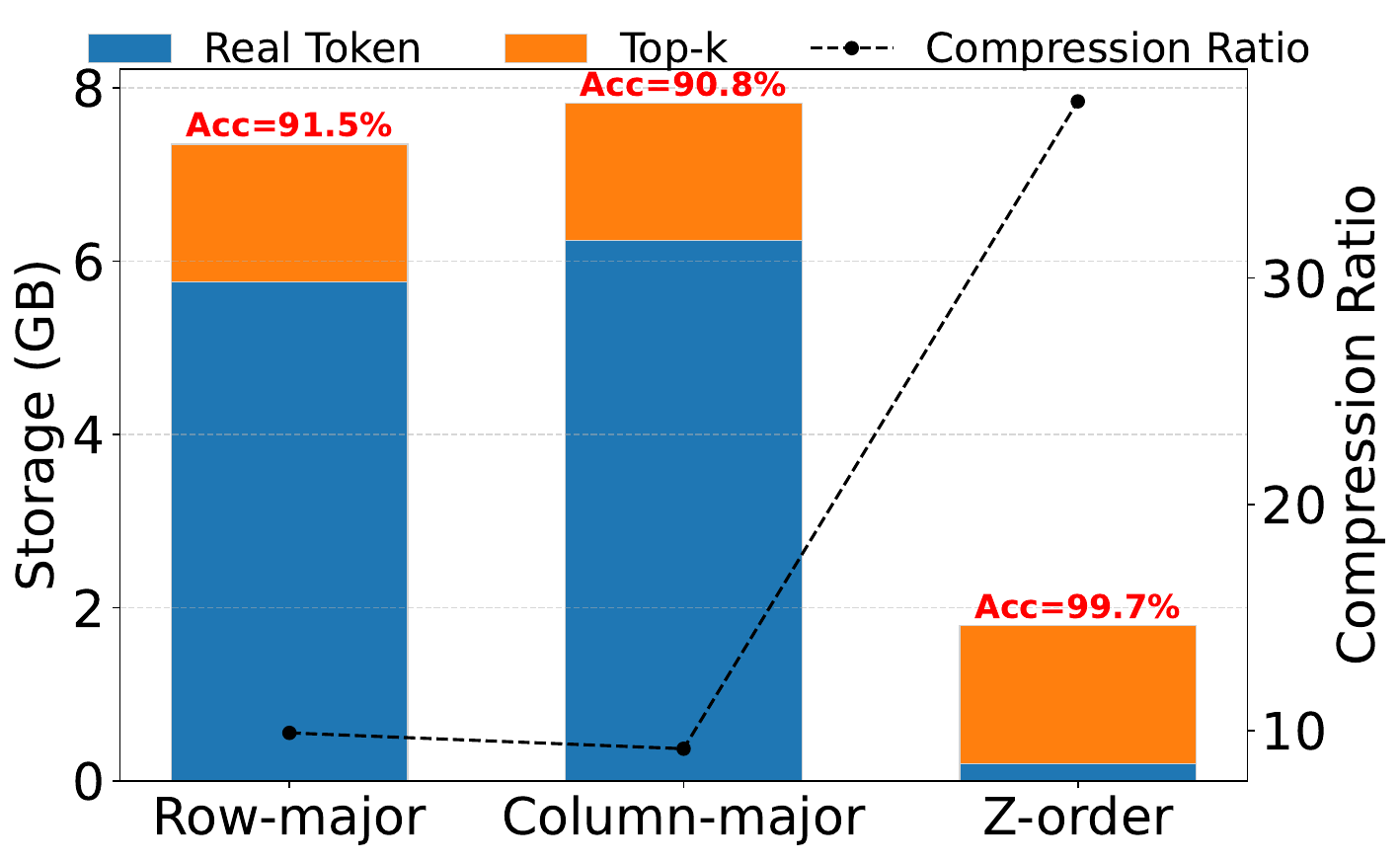}
    \end{subfigure}%
    \caption{Effect of data order on compression ratio and storage breakdown for \ERA5-2023-T ($\epsilon = 10^{-3}$).}
    \label{fig:exp:zorder-raw-column}
\end{figure}

\subsection{Effect of Data Order}\label{sec:exp:zorder_ablation}
We investigate how different data orders, namely row-major, column-major, and Z-order, affect compression performance under fixed vocabulary size 1024 and top-$k$ decoding ($k=8$). 
Figure~\ref{fig:exp:zorder-raw-column} presents the results on \ERA5-2023-T, reporting top-$k$ accuracy, overall storage (real token storage, top-$k$ index storage), and compression ratio.

Z-order achieves the best performance, attaining a compression ratio of $37.8$ and $99.7\%$ top-$k$ accuracy, compared to $\rho$ at $9.9$ (row-major) and $9.2$ (column-major). 
This improvement stems from the ability of Z-order to preserve spatial locality, which enhances sequential token predictability and reduces the need for fallback storage of real tokens.
In contrast, row-major and column-major traversals exhibit fragmented spatial patterns, leading to degraded predictability and larger fallback storage.
These findings highlight that spatial order is not merely a preprocessing choice but a critical factor for transformer-based compression models. 
Incorporating structure-aware ordering schemes like Z-order can significantly amplify the compression capability of autoregressive LLMs by exposing coherent patterns in the training sequence.

\subsection{Effect of Model Size}\label{sec:exp:model-size}
\begin{figure}[t]
    \centering
    \includegraphics[width=\linewidth]{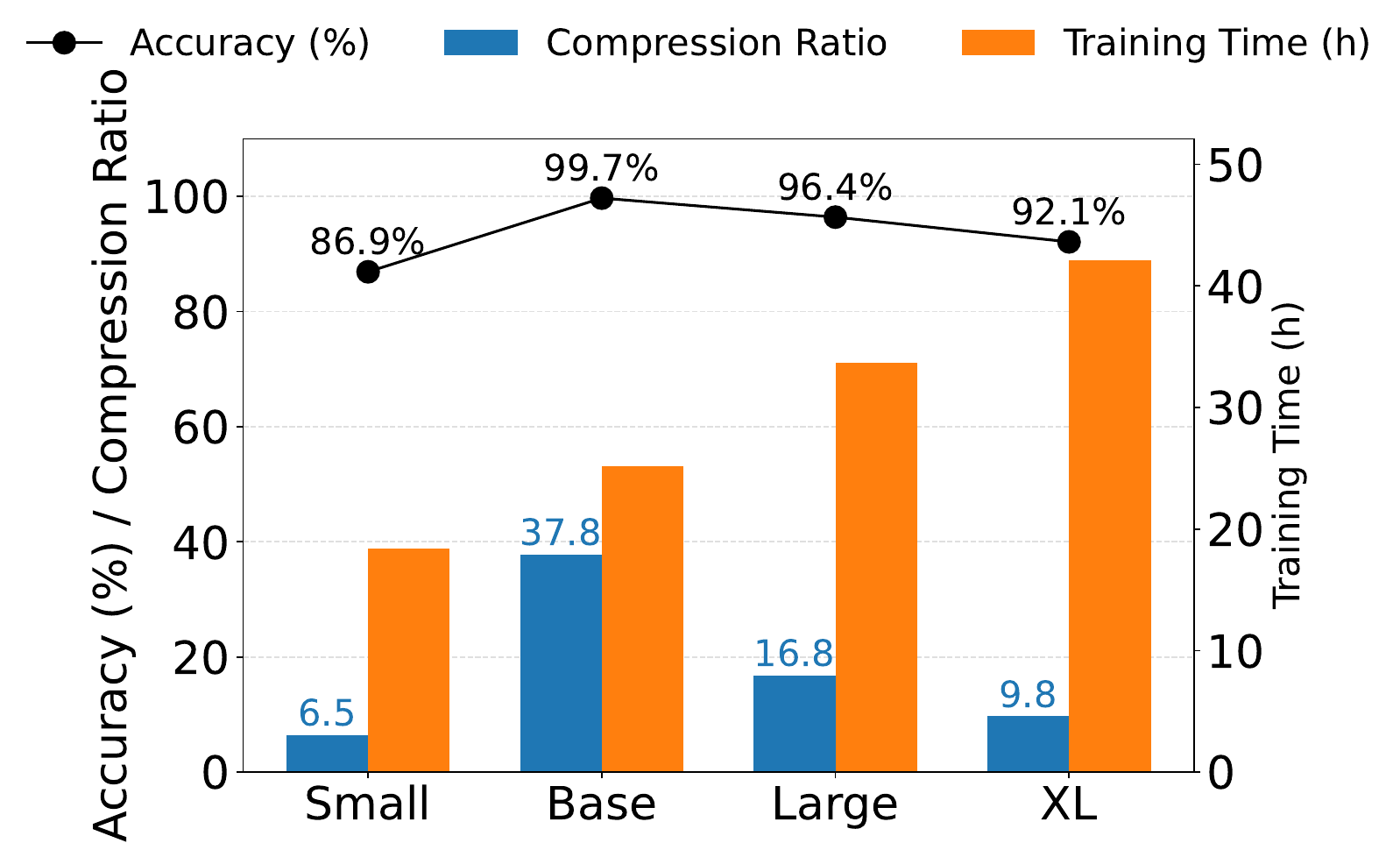}
    \caption{Effect of model size on accuracy, storage, and compression ratio on \ERA5-23-T ($k=8$, vocabulary size=1024, $\epsilon = 10^{-3}$).}
    \label{fig:exp:modelsize}
\end{figure}

We evaluate how the Transformer backbone size impacts compression performance using the \ERA5- cha23-T{} dataset, as shown in Figure~\ref{fig:exp:modelsize}.
We compare four GPT-style models, Small (117M), Base (345M), Large (762M), and XL (1.5B), trained on the same dataset for one epoch using 4$\times$A100 GPUs.

Interestingly, the Base model achieves the best compression ratio of 37.8$\times$, despite having only 345M parameters and moderate training cost (25 hours). This model also achieves the highest top-$k$ accuracy of 99.7\%, resulting in minimal fallback storage.
Increasing model size beyond this point does not yield consistent gains. The XL model (1.5B) requires 42 hours of training but achieves only 92.1\% accuracy and a lower compression ratio of 9.8$\times$. We hypothesize that over-parameterized models may overfit on frequent patterns while struggling to generalize rare transitions, leading to more prediction errors and higher correction cost.
On the other hand, the Small model underfits the training data, achieving only 86.9\% accuracy and resulting in large fallback storage (9.94GB). This suggests that a certain model capacity is necessary to capture the underlying structure of scientific data, but excessive scale may hurt generalization and compression.

Overall, our results demonstrate that model size must be carefully balanced  to achieve optimal compression performance under resource constraints.

\section{Related Work}\label{sec:Related Work}

\noindent{\bf Scientific data compression.}
High-resolution scientific datasets, from climate modeling~\cite{hoteit-RSRA2022} to seismic and fluid simulations~\cite{kayum2020geodrive}, pose significant storage and transmission challenges. Lossless compressors~\cite{cudre2009demonstration,zstandard,zlib} typically yield low ratios ($<2\times$)~\cite{zhao2020sdrbench}, making error-bounded lossy compression the dominant solution.
Transform-based methods like ZFP~\cite{lindstrom2014fixed} and SPERR~\cite{li2023lossy}, and predictive schemes in the SZ family~\cite{tao2017significantly,zhao2021optimizing}, combine curve fitting, Lorenzo predictors, neural-network predictors, and interpolation.
HPEZ~\cite{liu2023high} further integrates reordering, splines, and auto-tuned metrics to enhance fidelity and efficiency.

\noindent{\bf Transformers for scientific data.}
Transformer architectures, mostly based on encoder-decoder designs, have been widely applied beyond NLP~\cite{vaswani2017attention,devlin2018bert}, including in vision~\cite{liu2021Swin} and time series modeling~\cite{Zerveas-kdd2021,zuo2023svp,zuo2024darker}. Recent studies~\cite{jin2023time,gruver2023llmtime} explore adapting pretrained LLMs to numerical forecasting via prompt reprogramming or tokenization. However, the use of decoder-only transformers for structured scientific data remains an open direction.

However, despite these promising advances in time series modeling, no prior work has explored using LLMs for scientific data compression, where strict error-bounded reconstruction, long-range spatiotemporal dependencies, and symbolic efficiency are critical. 
Motivated by these gaps, we propose \LLMComp, a new framework that reformulates scientific data compression as a token-level sequential modeling task, leveraging decoder-only LLMs to achieve learnable, high-fidelity, error-bounded compression.

\section{Conclusion}\label{sec:conclusion}
In this paper, we present \LLMComp, a novel error-bounded lossy compression framework for scientific data based on decoder-only large language models (LLMs). 
By tokenizing scientific data and casting compression as a top-$k$ prediction task, \LLMComp~effectively captures rich spatiotemporal patterns while ensuring strict fidelity constraints. 
We further enhance performance via a coverage-guided sampling strategy that improves training efficiency, and adopt Z-order layouts to boost spatial locality and model accuracy. 
Extensive experiments across diverse climate and reanalysis datasets show that \LLMComp~consistently outperforms state-of-the-art compressors, achieving up to 30\% higher compression ratios under tight error bounds. 
These results highlight the potential of LLMs as general-purpose compressors for scientific data, offering strong adaptability and accuracy.
In the future, we aim to improve the training efficiency of \LLMComp~by exploring lightweight architectures, adaptive sampling, and curriculum learning strategies for large-scale scientific data.

\stab
\noindent{\bf Acknowledgments.}
For computer time, this research used IBEX and Shaheen III, managed by the Supercomputing Core Laboratory at King Abdullah University of Science and Technology (KAUST), Saudi Arabia.


\begin{thebibliography}{10}

\bibitem{achiam2023gpt}
Josh Achiam, Steven Adler, Sandhini Agarwal, Lama Ahmad, Ilge Akkaya,
  Florencia~Leoni Aleman, Diogo Almeida, Janko Altenschmidt, Sam Altman,
  Shyamal Anadkat, et~al.
\newblock Gpt-4 technical report.
\newblock {\em arXiv preprint}, 2023.

\bibitem{bai2023qwen}
Jinze Bai, Shuai Bai, Yunfei Chu, Zeyu Cui, Kai Dang, Xiaodong Deng, Yang Fan,
  Wenbin Ge, Yu~Han, Fei Huang, et~al.
\newblock Qwen technical report.
\newblock {\em arXiv preprint}, 2023.

\bibitem{cudre2009demonstration}
Philippe Cudr{\'e}-Mauroux, Hideaki Kimura, K-T Lim, Jennie Rogers, Roman
  Simakov, Emad Soroush, Pavel Velikhov, Daniel~L Wang, Magdalena Balazinska,
  Jacek Becla, et~al.
\newblock A demonstration of scidb: a science-oriented dbms.
\newblock {\em Proceedings of the VLDB Endowment}, 2(2):1534--1537, 2009.

\bibitem{hoteit-RSRA2022}
Hari~Prasad Dasari, Yesubabu Viswanadhapalli, Sabique Langodan, Yasser
  Abualnaja, Srinivas Desamsetti, Koteswararao Vankayalapati, Luong Thang, and
  Ibrahim Hoteit.
\newblock High-resolution climate characteristics of the arabian gulf based on
  a validated regional reanalysis.
\newblock {\em Meteorological Applications}, 29(5):e2102, 2022.

\bibitem{devlin2018bert}
Jacob Devlin, Ming{-}Wei Chang, Kenton Lee, and Kristina Toutanova.
\newblock {BERT:} pre-training of deep bidirectional transformers for language
  understanding.
\newblock In {\em Proceedings of the Nations of the Americas Chapter of the
  Association for Computational Linguistics (NAACL)}, 2019.

\bibitem{gray1998quantization}
Robert~M. Gray and David~L. Neuhoff.
\newblock Quantization.
\newblock {\em IEEE transactions on information theory}, 44(6):2325--2383,
  1998.

\bibitem{guo2025deepseek}
Daya Guo, Dejian Yang, Haowei Zhang, Junxiao Song, Ruoyu Zhang, Runxin Xu,
  Qihao Zhu, Shirong Ma, Peiyi Wang, Xiao Bi, et~al.
\newblock Deepseek-r1: Incentivizing reasoning capability in llms via
  reinforcement learning.
\newblock {\em arXiv preprint}, 2025.

\bibitem{han2024cra5}
Tao Han, Zhenghao Chen, Song Guo, Wanghan Xu, and Lei Bai.
\newblock Cra5: Extreme compression of era5 for portable global climate and
  weather research via an efficient variational transformer.
\newblock {\em arXiv preprint}, 2024.

\bibitem{hersbach2020era5}
Hans Hersbach, Bill Bell, Paul Berrisford, Shoji Hirahara, Andr{\'a}s
  Hor{\'a}nyi, Joaqu{\'\i}n Mu{\~n}oz-Sabater, Julien Nicolas, Carole Peubey,
  Raluca Radu, Dinand Schepers, et~al.
\newblock The era5 global reanalysis.
\newblock {\em Quarterly Journal of the Royal Meteorological Society},
  146(730):1999--2049, 2020.

\bibitem{huang2023compressing}
Langwen Huang and Torsten Hoefler.
\newblock Compressing multidimensional weather and climate data into neural
  networks.
\newblock In {\em Proceedings of the International Conference on Learning
  Representations (ICLR)}, 2023.

\bibitem{jin2023time}
Ming Jin, Shiyu Wang, Lintao Ma, Zhixuan Chu, James~Y Zhang, Xiaoming Shi,
  Pin-Yu Chen, Yuxuan Liang, Yuan-Fang Li, Shirui Pan, and Qingsong Wen.
\newblock {Time-LLM}: Time series forecasting by reprogramming large language
  models.
\newblock In {\em Proceedings of the International Conference on Learning
  Representations (ICLR)}, 2024.

\bibitem{kay2015community}
Jennifer~E Kay, Clara Deser, A~Phillips, A~Mai, Cecile Hannay, Gary Strand,
  Julie~Michelle Arblaster, SC~Bates, Gokhan Danabasoglu, James Edwards, et~al.
\newblock The community earth system model (cesm) large ensemble project: a
  community resource for studying climate change in the presence of internal
  climate variability.
\newblock {\em Bulletin of the American Meteorological Society},
  96(8):1333--1349, 2015.

\bibitem{kayum2020geodrive}
Suha~N Kayum, Thierry Tonellot, Vincent Etienne, Ali Momin, Ghada Sindi, Maxim
  Dmitriev, and Hussain Salim.
\newblock Geodrive-a high performance computing flexible platform for seismic
  applications.
\newblock {\em First Break}, 38(2):97--100, 2020.

\bibitem{lam2023learning}
Remi Lam, Alvaro Sanchez-Gonzalez, Matthew Willson, Peter Wirnsberger, Meire
  Fortunato, Ferran Alet, Suman Ravuri, Timo Ewalds, Zach Eaton-Rosen, Weihua
  Hu, et~al.
\newblock Learning skillful medium-range global weather forecasting.
\newblock {\em Science}, 382(6677):1416--1421, 2023.

\bibitem{hoteit-RSRA2018}
Sabique Langodan, Luigi Cavaleri, Angela Pomaro, Jesus Portilla, Yasser
  Abualnaja, and Ibrahim Hoteit.
\newblock Unraveling climatic wind and wave trends in the red sea using wave
  spectra partitioning.
\newblock {\em Journal of Climate}, 31(5):1881 -- 1895, 2018.

\bibitem{li2025graphcomp}
Guozhong Li, Muhannad Alhumaidi, Spiros Skiadopoulos, Ibrahim Hoteit, and Panos
  Kalnis.
\newblock Graphcomp: Extreme error-bounded compression of scientific data via
  temporal graph autoencoders.
\newblock {\em arXiv preprint}, 2025.

\bibitem{li2025llmcomp}
Guozhong Li, Muhannad Alhumaidi, Spiros Skiadopoulos, and Panos Kalnis.
\newblock Llmcomp: A language modeling paradigm for error-bounded scientific
  data compression.
\newblock {\em arXiv preprint}, 2025.

\bibitem{li2025lesax}
Guozhong Li, Byron Choi, Rundong Zuo, Sourav~S Bhowmick, and Jianliang Xu.
\newblock lesax index: A learned sax representation index for time series
  similarity search.
\newblock In {\em Proceedings of the IEEE International Conference on Data
  Engineering (ICDE)}, pages 1995--2008. IEEE Computer Society, 2025.

\bibitem{li2023lossy}
Shaomeng Li, Peter Lindstrom, and John Clyne.
\newblock Lossy scientific data compression with sperr.
\newblock In {\em Proceedings of the IEEE International Parallel and
  Distributed Processing Symposium (IPDPS)}, pages 1007--1017, 2023.

\bibitem{lindstrom2014fixed}
Peter Lindstrom.
\newblock Fixed-rate compressed floating-point arrays.
\newblock {\em IEEE TVCG}, 20(12):2674--2683, 2014.

\bibitem{liu2023srn}
Jinyang Liu, Sheng Di, Sian Jin, Kai Zhao, Xin Liang, Zizhong Chen, and Franck
  Cappello.
\newblock Srn-sz: scientific error-bounded lossy compression with
  super-resolution neural networks.
\newblock In {\em Proceedings of the IEEE international conference on Big Data
  (Big Data)}, pages 229--236, 2023.

\bibitem{liu2021exploring}
Jinyang Liu, Sheng Di, Kai Zhao, Sian Jin, Dingwen Tao, Xin Liang, Zizhong
  Chen, and Franck Cappello.
\newblock Exploring autoencoder-based error-bounded compression for scientific
  data.
\newblock In {\em Proceedings of the IEEE International Conference on Cluster
  Computing (CLUSTER)}, pages 294--306, 2021.

\bibitem{liu2023high}
Jinyang Liu, Sheng Di, Kai Zhao, Xin Liang, Sian Jin, Zizhe Jian, Jiajun Huang,
  Shixun Wu, Zizhong Chen, and Franck Cappello.
\newblock High-performance effective scientific error-bounded lossy compression
  with auto-tuned multi-component interpolation.
\newblock {\em SIGMOD}, 2(1):1--27, 2024.

\bibitem{liu2021Swin}
Ze~Liu, Yutong Lin, Yue Cao, Han Hu, Yixuan Wei, Zheng Zhang, Stephen Lin, and
  Baining Guo.
\newblock Swin transformer: Hierarchical vision transformer using shifted
  windows.
\newblock In {\em Proceedings of the International Conference on Computer
  Vision (ICCV)}, 2021.

\bibitem{mirowski2024neural}
Piotr Mirowski, David Warde-Farley, Mihaela Rosca, Matthew~Koichi Grimes, Yana
  Hasson, Hyunjik Kim, M{\~A}{\v{S}}lanie Rey, Simon Osindero, Suman Ravuri,
  and Shakir Mohamed.
\newblock Neural compression of atmospheric states.
\newblock {\em arXiv preprint}, 2024.

\bibitem{gruver2023llmtime}
Shikai~Qiu Nate~Gruver, Marc~Finzi and Andrew~Gordon Wilson.
\newblock {Large Language Models Are Zero Shot Time Series Forecasters}.
\newblock In {\em Proceedings of Neural Information Processing Systems
  (NeurIPS)}, 2023.

\bibitem{pearlman2004efficient}
William~A Pearlman, Asad Islam, Nithin Nagaraj, and Amir Said.
\newblock Efficient, low-complexity image coding with a set-partitioning
  embedded block coder.
\newblock {\em IEEE TCSVT}, 14(11):1219--1235, 2004.

\bibitem{price2025probabilistic}
Ilan Price, Alvaro Sanchez-Gonzalez, Ferran Alet, Tom~R Andersson, Andrew
  El-Kadi, Dominic Masters, Timo Ewalds, Jacklynn Stott, Shakir Mohamed, Peter
  Battaglia, et~al.
\newblock Probabilistic weather forecasting with machine learning.
\newblock {\em Nature}, 637(8044):84--90, 2025.

\bibitem{radford2018improving}
Alec Radford, Karthik Narasimhan, Tim Salimans, Ilya Sutskever, et~al.
\newblock Improving language understanding by generative pre-training.
\newblock 2018.

\bibitem{tang2006three}
Xiaoli Tang and William~A Pearlman.
\newblock Three-dimensional wavelet-based compression of hyperspectral images.
\newblock In {\em Hyperspectral data compression}, pages 273--308. Springer,
  2006.

\bibitem{tao2017significantly}
Dingwen Tao, Sheng Di, Zizhong Chen, and Franck Cappello.
\newblock Significantly improving lossy compression for scientific data sets
  based on multidimensional prediction and error-controlled quantization.
\newblock In {\em Proceedings of the IEEE International Parallel and
  Distributed Processing Symposium (IPDPS)}, pages 1129--1139, 2017.

\bibitem{touvron2023llama}
Hugo Touvron, Thibaut Lavril, Gautier Izacard, Xavier Martinet, Marie-Anne
  Lachaux, Timoth{\'e}e Lacroix, Baptiste Rozi{\`e}re, Naman Goyal, Eric
  Hambro, Faisal Azhar, et~al.
\newblock Llama: Open and efficient foundation language models.
\newblock {\em arXiv preprint}, 2023.

\bibitem{vaswani2017attention}
Ashish Vaswani, Noam Shazeer, Niki Parmar, Jakob Uszkoreit, Llion Jones,
  Aidan~N Gomez, {\L}ukasz Kaiser, and Illia Polosukhin.
\newblock Attention is all you need.
\newblock In {\em Proceedings of Neural Information Processing Systems
  (NeurIPS)}, 2017.

\bibitem{yan2022sensor}
Li~Yan, Nerissa Xu, Guozhong Li, Sourav~S Bhowmick, Byron Choi, and Jianliang
  Xu.
\newblock Sensor: data-driven construction of sketch-based visual query
  interfaces for time series data.
\newblock {\em VLDB}, 15(12):3650--3653, 2022.

\bibitem{Zerveas-kdd2021}
George Zerveas, Srideepika Jayaraman, Dhaval Patel, Anuradha Bhamidipaty, and
  Carsten Eickhoff.
\newblock A transformer-based framework for multivariate time series
  representation learning.
\newblock In {\em Proceedings of the ACM International Conference on Knowledge
  discovery and Data Mining (SIGKDD)}, page 2114–2124, 2021.

\bibitem{zhao2021optimizing}
Kai Zhao, Sheng Di, Maxim Dmitriev, Thierry-Laurent~D Tonellot, Zizhong Chen,
  and Franck Cappello.
\newblock Optimizing error-bounded lossy compression for scientific data by
  dynamic spline interpolation.
\newblock In {\em Proceedings of the IEEE International Conference on Data
  Engineering (ICDE)}, pages 1643--1654, 2021.

\bibitem{zhao2020sdrbench}
Kai Zhao, Sheng Di, Xin Lian, Sihuan Li, Dingwen Tao, Julie Bessac, Zizhong
  Chen, and Franck Cappello.
\newblock Sdrbench: scientific data reduction benchmark for lossy compressors.
\newblock In {\em Proceedings of the IEEE international conference on Big Data
  (Big Data)}, pages 2716--2724, 2020.

\bibitem{zlib}
{Zlib}.
\newblock Zlib.
\newblock \url{http://www.zlib.net/}.
\newblock Accessed: Jan. 22, 2024.

\bibitem{zstandard}
{Zstandard}.
\newblock Zstandard.
\newblock \url{www.zstd.net}.
\newblock Accessed: Feb. 16, 2024.

\bibitem{zuo2024darker}
Rundong Zuo, Guozhong Li, Rui Cao, Byron Choi, Jianliang Xu, and Sourav~S
  Bhowmick.
\newblock Darker: Efficient transformer with data-driven attention mechanism
  for time series.
\newblock {\em Proceedings of the VLDB Endowment}, 17(11):3229--3242, 2024.

\bibitem{zuo2023svp}
Rundong Zuo, Guozhong Li, Byron Choi, Sourav~S Bhowmick, Daphne Ngar-yin Mah,
  and Grace~LH Wong.
\newblock Svp-t: A shape-level variable-position transformer for multivariate
  time series classification.
\newblock In {\em Annual AAAI Conference on Artificial Intelligence (AAAI)},
  volume~37, pages 11497--11505, 2023.

\end{thebibliography}
\end{document}